\documentclass[5p,nopreprintline]{elsarticle}
\newcommand{\variable}{\kappa}%
\newcommand{\numberVar}{m}

\usepackage{amsmath,amsfonts}
\usepackage{algorithmic}
\usepackage{algorithm}
\usepackage{array}
\usepackage{booktabs}
\usepackage[caption=false,font=normalsize,labelfont=sf,textfont=sf]{subfig}
\usepackage{textcomp}
\usepackage{stfloats}
\usepackage{url}
\usepackage{verbatim}
\usepackage{graphicx}
\usepackage[font=small,labelfont=bf,tableposition=top]{caption}
\usepackage{tikz}
\usepackage[short]{optidef}
\usetikzlibrary{chains}
\usetikzlibrary{calc}
\usetikzlibrary{fit}

\usepackage{theorem, amsfonts, latexsym, amscd, amsopn, amssymb, amsmath, bbm, stmaryrd, wasysym, mathbbol, calrsfs}
\usepackage{color}
\usepackage{microtype}      
\usepackage{todonotes}
\usepackage{hyperref}
\hypersetup{
	colorlinks,
	linkcolor={red!80!black},
	citecolor={blue!80!black},
	urlcolor={blue!80!black}
}
\usepackage{cleveref} 
\usepackage{autonum} 
\usepackage{glossaries-extra}
\usepackage{subfig}
\usepackage{siunitx}

\usepackage{pgfplots}
\pgfplotsset{compat=newest}
\pgfplotsset{scaled y ticks=false}
\usepgfplotslibrary{groupplots}
\usepgfplotslibrary{dateplot}

\usepackage{enumitem}
\newlist{todolist}{itemize}{2}
\setlist[todolist]{label=$\square$}
\usepackage{pifont}

\GlsXtrEnableEntryCounting
{abbreviation}
{3}

\newabbreviation{AI}{AI}{Artificial Intelligence}
\newabbreviation{AO}{AO}{All-Optical}
\newabbreviation{AWGN}{AWGN}{Additive White Gaussian Noise}
\newabbreviation{BP}{BP}{Back--Propagation}
\newabbreviation{DL}{DL}{Deep Learning}
\newabbreviation{InP}{InP}{Indium phosphide}
\newabbreviation{MAC}{MAC}{Multiply-–Accumulate Operation}
\newabbreviation{ML}{ML}{Machine Learning}
\newabbreviation{MNIST}{MNIST}{Modified National Institute of Standards and Technology}
\newabbreviation{MSE}{MSE}{Mean Squared Error}
\newabbreviation{MRR}{MRR}{Micro--Ring--Resonator}
\newabbreviation{MZI}{MZI}{Mach--Zehnder Interferometer}
\newabbreviation{NN}{NN}{Neural Network}
\newabbreviation{OIU}{OIU}{Optical Interference Unit}
\newabbreviation{ONN}{ONN}{Optical Neural Network}
\newabbreviation{O/E/O}{O/E/O}{optical/electrical/optical}
\newabbreviation{SGD}{SGD}{Stochastic Gradient Descent}
\newabbreviation{SOA}{SOA}{Semiconductor Optical Amplifier}
\newabbreviation{WDM}{WDM}{Wavelength--Division Multiplexing}

\newlength\myheight
\newlength\mydepth
\settototalheight\myheight{Xygp}
\newcommand*\inlinegraphics[1]{%
	\settototalheight\myheight{Xygp}%
	\settodepth\mydepth{Xygp}%
	\raisebox{-\mydepth}{\includegraphics[height=\myheight]{#1}}%
}

\newcommand{\R}{\mathbb{R}}

\newtheorem{corollary}{Corollary}

\newtheorem{theorem}{Theorem}
\newtheorem{proposition}{Proposition}

\allowdisplaybreaks

\usepackage{mathtools}
\usepackage{etoolbox}

\usepackage{dsfont} 

\newcommand{\pnorm}[2]{ \| #1 \|{}_{#2} }

\newcommand{\naturalNumbersPlus}{ \mathbb{N}_{+} }

\newcommand{\QuodEratDemonstrandum}{\hfill \ensuremath{\Box}}

\def\eqcom#1{\overset{\textnormal{(#1)}}}

\def\({{\Bigl(}}
        \def\){{\Bigr)}}

\newcommand{\ba}{\begin{array}}
    \newcommand{\ea}{\end{array}}

\newcommand{\xdeleted}[1]{\deleted{}} 

\begin{document}

\begin{frontmatter}
	\title{Noise-Resilient Designs for Optical Neural Networks}
	\author[1,2]{Gianluca Kosmella\corref{cor}}
	\cortext[cor]{Corresponding author}
	\ead{g.k.kosmella@tue.nl}
	\author[1]{Ripalta Stabile}
	\author[2]{Jaron Sanders}

	\affiliation[1]{organization={Department of Electrical Engineering, Eindhoven University of Technology},
		addressline={PO Box 513},
		postcode={5600 MB},
		postcodesep={},
		city={Eindhoven},
		country={The Netherlands}}

	\affiliation[2]{organization={Department of Mathematics and Computer Science, Eindhoven University of Technology},
		addressline={PO Box 513},
		postcode={5600 MB},
		postcodesep={},
		city={Eindhoven},
		country={The Netherlands}}

	\begin{abstract}
		All analog signal processing is fundamentally subject to noise, and this is also the case in modern implementations of \glspl{ONN}.
Therefore, to mitigate noise in \glspl{ONN}, we propose two designs
that are constructed from a given, possibly trained, \gls{NN} that one wishes to implement.
Both designs have the capability that the resulting \glspl{ONN} gives outputs close to the desired \gls{NN}.

To establish the latter, we analyze the designs mathematically.
Specifically, we investigate a probabilistic framework for the first design that establishes that the design is correct, i.e., for any feed-forward \gls{NN} with Lipschitz continuous activation functions, an \gls{ONN} can be constructed that produces output arbitrarily close to the original.
\glspl{ONN} constructed with the first design thus also inherit the universal approximation property of \glspl{NN}.
For the second design, we restrict the analysis to \glspl{NN} with linear activation functions and characterize the \glspl{ONN}' output distribution using exact formulas.

Finally, we report on numerical experiments with LeNet \glspl{ONN} that give insight into the number of components required in these designs for certain accuracy gains.
We specifically study the effect of noise as a function of the depth of an \gls{ONN}.
The results indicate that in practice, adding just a few components in the manner of the first or the second design can already be expected to increase the accuracy of \glspl{ONN} considerably.

	\end{abstract}

\end{frontmatter}

\section{Introduction}
\glsreset{ONN}
\glsreset{NN}

\gls{ML} is a computing paradigm in which problems that are traditionally challenging for programmers to explicitly write algorithms for, are solved by learning algorithms that improve automatically through experience.
That is, they ``learn'' structure in data.
Prominent examples include image recognition \cite{he2016deep}, semantic segmentation \cite{long2015fully}, human-level control in video games \cite{mnih2015human}, visual tracking \cite{nam2016learning}, and language translation \cite{wu2016google}.

Classical computers are designed and best suited for serialized operations (they have a central processing unit and separated memory), while the data-driven \gls{ML} approach requires decentralized and parallel calculations at high bandwidth as well as continuous processing of parallel data.
To illustrate how \gls{ML} can benefit from a different architecture, we can consider performance relative to the number of executed operations, also indicated as \gls{MAC} rates, and the energy efficiency, i.e., the amount of energy spent to execute one single operation.
Computational efficiency in classical computers levels off below $10$ G\gls{MAC}/s/W \cite{hasler2013finding}.

An alternative computing architecture with a more distributed interconnectivity and memory would allow for greater energy efficiency and computational speed. An inspiring example would be an architecture such as the brain.
The brain is able to perform about $10^{18}$ \gls{MAC}/s using only $\SI{20}{\watt}$ of power \cite{hasler2013finding}, and operates approximately $10^{11}$ neurons with an average number of inputs for each of about $10^4$ synapses.
This leads to an estimated total of $10^{15}$ synaptic connections, all conveying signals up to $\SI{1}{\kilo \hertz}$ bandwidth.
The brain's computational efficiency (being less than \SI{1}{\atto\joule} per \gls{MAC}) is then about $8$ orders of magnitude higher than the one of current supercomputers, which operate instead at $\SI{100}{\pico \joule}$ per \gls{MAC} \cite{hasler2013finding}.

Connecting software to hardware through computing architecture tailored to \gls{ML} tasks is the endeavor of research within the field of neuromorphic computing.
The electronics community is now busy developing non-von Neumann computing architectures to enable information processing with an energy efficiency down to a few pJ per operation.
Aiming to replicate fundamentals of biological neural circuits in dedicated hardware, important advances have been made in neuromorphic accelerators \cite{du2015neuromorphic}.
These advances are based on the spiking architectural models, which are still not fully understood.
\gls{DL}-focused approaches, on the other hand, aim to construct hardware that efficiently realizes \gls{DL} architectures, while eliminating as much of the complexity of biological neural networks as possible.
Among the most powerful \gls{DL} hardware we can name the GPU-based \gls{DL} accelerators hardware \cite{akopyan2015truenorth,benjamin2014neurogrid,furber2014spinnaker,merolla2014millionTRUENORTH,schemmel2010waferHICANN}, as well as emerging analogue electronic \gls{AI} chipsets that tend to collocate processing and memory to minimize the memory–processor communication energy costs (e.g.\ the analogue crossbar approaches \cite{siddiqui2019magnetic}).
The {\href{https://mythic.ai/}{Mythic}'}s architecture, for example, can yield high accuracy in inference applications within a remarkable energy efficiency of just half a pJ per \gls{MAC}.
Even if the implementation of neuromorphic approaches is visibly bringing outstanding record energy efficiencies and computation speeds, neuromorphic electronics is already struggling to offer the desired data throughput at the neuron level.
Neuromorphic processing for high-bandwidth applications requires GHz operation per neuron, which calls for a fundamentally different technology approach.

\subsection{Optical Neural Networks}

A major concern with neuromorphic electronics is that the distributed hardware needed for parallel interconnections is impractical to realize with classical metal wiring: a trade-off applies between interconnectivity and bandwidth, limiting these engine's utilization to applications in the kHz and sub-GHz regime.
When sending information not through electrical signals but via optical signals, the optical interconnections do not undergo interference and the optical bandwidth is virtually unlimited.
This can for example be achieved when exploiting the color and/or the space and/or the polarization and/or the time domain, thus allowing for applications in the GHz regime.
It has been theorized that photonic neuromorphic processors could operate ten thousand times faster while using less energy per computation \cite{de2017progress, kitayama2019novel, miscuglio2020artificial, shastri2017principles}.
Photonics therefore seems to be a promising platform for advances in neuromorphic computing.

Implementations of weighted addition for \glspl{ONN} include \gls{MZI}-based \glspl{OIU} \cite{shen2017deep}, time-multiplexed and, coherent detection \cite{hamerly2019large}, free space systems using spatial light modulators \cite{bernstein2021freely}
and \gls{MRR}-based weighting bank on silicone \cite{huang2020demonstration}.
Furthermore, \gls{InP}-integrated optical cross-connect using \glspl{SOA} as single stage weight elements, as well as \gls{SOA}-based wavelength converters \cite{shi2019deep, shi2020lossless, shi2019first} have been demonstrated for allowing \gls{AO} \glspl{NN}.
A comprehensive review of all the approaches used in integrated photonics can be found in \cite{shastri2021photonics}.

Next to these promises, aspects like implementation of nonlinearities, access and storage of weights in on-chip memory, and noise sources in analog photonic implementations,
all pose challenges in devising scalable photonic neuromorphic processors and accelerators.
These challenges also occur when they are embedded within end-to-end systems.
Fortunately, arbitrary scalability of these networks has been demonstrated, with a certain noise and accuracy.
However, it would be useful to envision new architectures to reduce noise even more.

\subsection{Noise in \texorpdfstring{\glspl{ONN}}{ONNs}}
\label{sec:Introduction__Modeling_noise_in_ONNs}

The types of noise in \glspl{ONN} include thermal crosstalk \cite{tait2017neuromorphic}, cumulative noise in optical communication links \cite{essiambre2010capacity, li2007channel} and noise deriving from applying an activation function \cite{de2019noise}.

In all these studies, the noise is considered to be approximated well by \gls{AWGN}.

For example, taking the studies \cite{tait2017neuromorphic, li2007channel, essiambre2010capacity, de2019noise, chakraborty2018toward} as starting point, the authors of \cite{passalis2021training} model an \gls{ONN} as a communication channel with \gls{AWGN}.
We follow this assumption and will model an \gls{ONN} as having been built up from interconnected nodes with noise in between them. This generic approach does not restrict us to any specific device that may be used in practice.

The model also applies to the two alternative designs of an \gls{AO} implementation of a \gls{NN} (see for example \cite{allopticalsigmoidmourgias2019}) and the case of an \gls{O/E/O} \gls{NN} \cite{shi2019deep}.
In an \gls{AO} \gls{NN}, the activation function is applied by manipulating an incoming electromagnetic wave.
Modulation (and the \gls{AWGN} it causes) only occurs prior to entering an \gls{AO}\ \gls{NN} (or equivalently, in the first layer).
For the remainder of the network the signal remains in the optical domain.
Here, when applying the optical activation function a new source of noise is introduced as \gls{AWGN} at the end of each layer.
Using the \gls{O/E/O} network architecture, the weighted addition is performed in the optical realm, but the light is captured soon after each layer, where it is converted into an electrical and digital signal and the activation function is applied via software on a computer.
The operation on the computer can be assumed to be noiseless.
However, since the result again needs to be modulated (to be able to act as input to the next layer), modulation noise is added.
We can further abstract from the specifics of the \gls{AO} and \gls{O/E/O} design and see that in either implementation noise occurs at the same locations within the mathematical modeling, namely \gls{AWGN} for weighted addition and afterwards \gls{AWGN} from an optical activation function or from modulation, respectively.
This means that we do not need to distinguish between the two design choices in our modeling; we only need to choose the corresponding \gls{AWGN} term after activation.

The operation of a layer of a feed-forward \gls{NN} can be modeled by multiplying a matrix $W$ with an input vector $x$ (a bias term $b$ can be absorbed into the matrix--vector product and will therefore suppressed in notation here) and then applying an activation function $f:\mathbb{R} \rightarrow \mathbb{R}$ element-wise to the result.
Symbolically,
\begin{align}
	x \mapsto f(Wx).
\end{align}

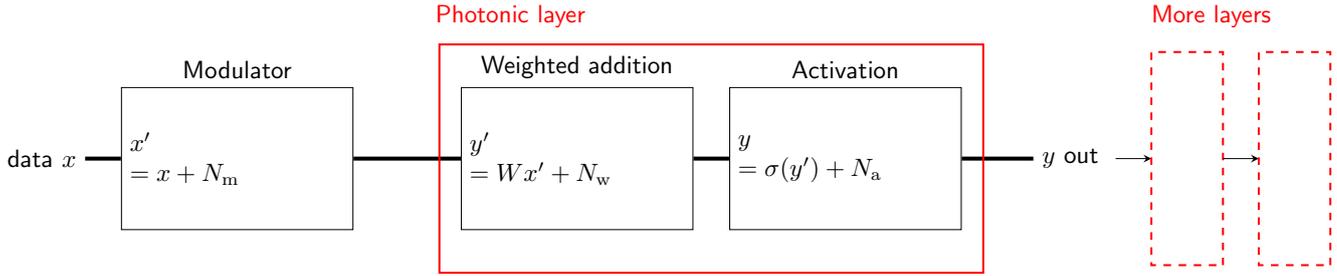
\begin{figure*}[hbtp]
	\centering
	\resizebox{1.05\textwidth}{!}{%
			\begin{tikzpicture}[
	start chain=going right,
	box/.style={
		on chain,join,draw,
		minimum height=3cm,
		text centered,
		minimum width=2cm,
	},
	every join/.style={ultra thick},
	node distance=5mm
	]
	\node [on chain] {data $x$}; 
	
	\node [on chain,join,draw,
	xshift=0mm,
	text width=3cm,
	minimum width=3cm,
	minimum height=2cm,
	label=above:Modulator,
	] (mod) {
		$x^\prime$\\
		$= x+N_\mathrm{m}$
	};
	
	\node [on chain,join,draw, 
	xshift=10mm,
	text width=3cm,
	minimum width=3cm,
	minimum height=2cm,
	label=above:Weighted addition,
	] (wei_add) {
		$y^\prime$\\
		$= Wx^\prime + N_\mathrm{w}$
	};
	
	\node [on chain,join,draw, 
	text width=3cm,
	minimum width=3cm,
	minimum height=2cm,
	label=above:Activation,
	] (act) {
		$y$\\
		$= \sigma (y^\prime) + N_\mathrm{a}$
	};
	
	\node [on chain,join,xshift=5mm]{$y$ out};

	\draw[red,thick] ($(wei_add.north west)+(-0.3,0.6)$)  rectangle ($(act.south east)+(0.3,-0.6)$);
	\node[red,text width=4cm] at (7.5,2) {Photonic layer};
	
	\draw [-stealth](15,0) -- (15.5,0);
	
	\draw[red,thick,dashed] (16.5,-1.5)  rectangle (15.5,1.5);
	
	\draw [-stealth](16.5,0) -- (17,0);
	
	\draw[red,thick,dashed] (18,-1.5)  rectangle (17,1.5);
	
	\node[red,text width=4cm] at (17.5,2) {More layers};
	
\end{tikzpicture}
	}%
	\caption{
		Schematic depiction of the noise model of \glspl{ONN} that we study. First, data $x$ is modulated onto light.
		This step adds an \gls{AWGN} term $N_\mathrm{m}$.
		This light enters the Photonic Layer, in which a weighted addition takes place, adding \gls{AWGN} $N_\mathrm{w}$.
		The activation function is then applied, adding \gls{AWGN} $N_\mathrm{a}$.
		The activation function may be applied by photo-detecting the signal of the weighted addition, turning it to a digital signal and applying the activation function on a computer.
		The result of that action would then be modulated again, to produce the optical output of the photonic neuron.
		The modulator is thus only required in the first layer, as each photonic neuron takes in light and outputs light.
	}
	\label{fig: possible_fig_for_modelling_section}
\end{figure*}

Now, concretely, the noise model that we study is described by
\begin{align}
	x \mapsto f({Wx} + \mathrm{Normal}(0, \Sigma_{\mathrm{w}} ) )+ \mathrm{Normal}(0, \Sigma_{\mathrm{a}} ),\label{noise description}
\end{align}
for each hidden layer of the \gls{ONN}.
Here $\mathrm{Normal}(0, \Sigma )$ denotes the multivariate normal distribution with mean vector ${0}$ and covariance matrix $\Sigma$.
More specifically, $\Sigma_{\mathrm{w}}$, $\Sigma_{\mathrm{a}}$ and $\Sigma_{\mathrm{m}}$ are the covariance matrices associated with weighted addition, application of the activation function, and modulation, respectively. \Cref{fig: possible_fig_for_modelling_section} gives a schematic representation of the noise model under study.
As we have seen above, in the \gls{O/E/O} case we have $\Sigma_{\mathrm{a}}=\Sigma_{\mathrm{m}}$, otherwise $\Sigma_{\mathrm{a}}$ is due to the specific structure of the photonic activation function.
The first layer, regardless of an \gls{AO} or \gls{O/E/O} network, sees a modulated input $x$, i.e., $x+\mathrm{Normal}({0}, \Sigma_{\mathrm{m}})$, and afterwards the same steps of weighing and applying an activation function, that is \eqref{noise description}.
Arguably the hidden layers and their noise structure are the most important parts, especially in deep \glspl{NN}.
Therefore, the main equation governing the behavior of the noise propagation in an \gls{ONN} will remain \eqref{noise description}.

\subsection{Noise-resistant designs for \texorpdfstring{\glspl{ONN}}{ONNs}}
\label{sec:Introduction__Designing_noise-resiliant_ONNs}

The main contribution of this paper lies in analyzing two noise reduction mechanisms for feed-forward \glspl{ONN}.
The mechanisms are derived from the insight that noise can be mitigated through averaging because of the law of large numbers, and they are aimed at using the enormous bandwidth that photonics offer.
The first design (Design A) and its analysis are inspired by recent advancements for \glspl{NN} with random edges in \cite{manita2020universal};
the second design (Design B) is new and simpler to implement, but comes without a theoretical guarantee of correctness for nonlinear \glspl{ONN}, specifically.

\begin{figure}[hbtp]
	\subfloat[The original \gls{NN}.]{
		\centering
		\includegraphics[width=0.975\linewidth]{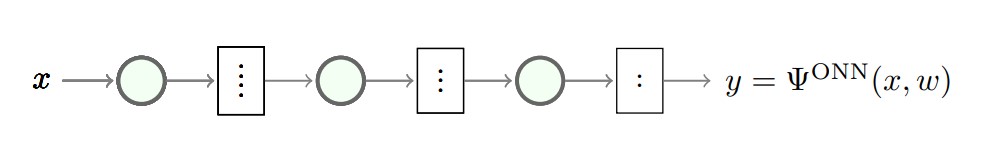}
	}
	\\
	\subfloat[Design A.]{
		\centering
		\includegraphics[width=0.975\linewidth]{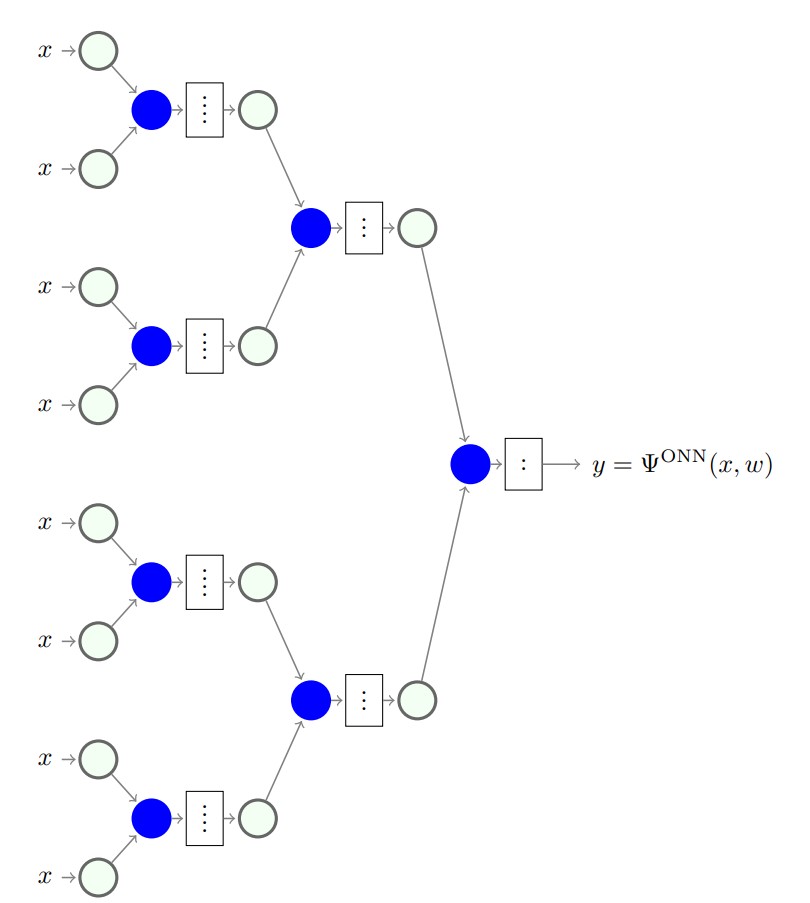}
	}
	\\
	\subfloat[Design B.]{
		\centering
		\includegraphics[width=0.975\linewidth]{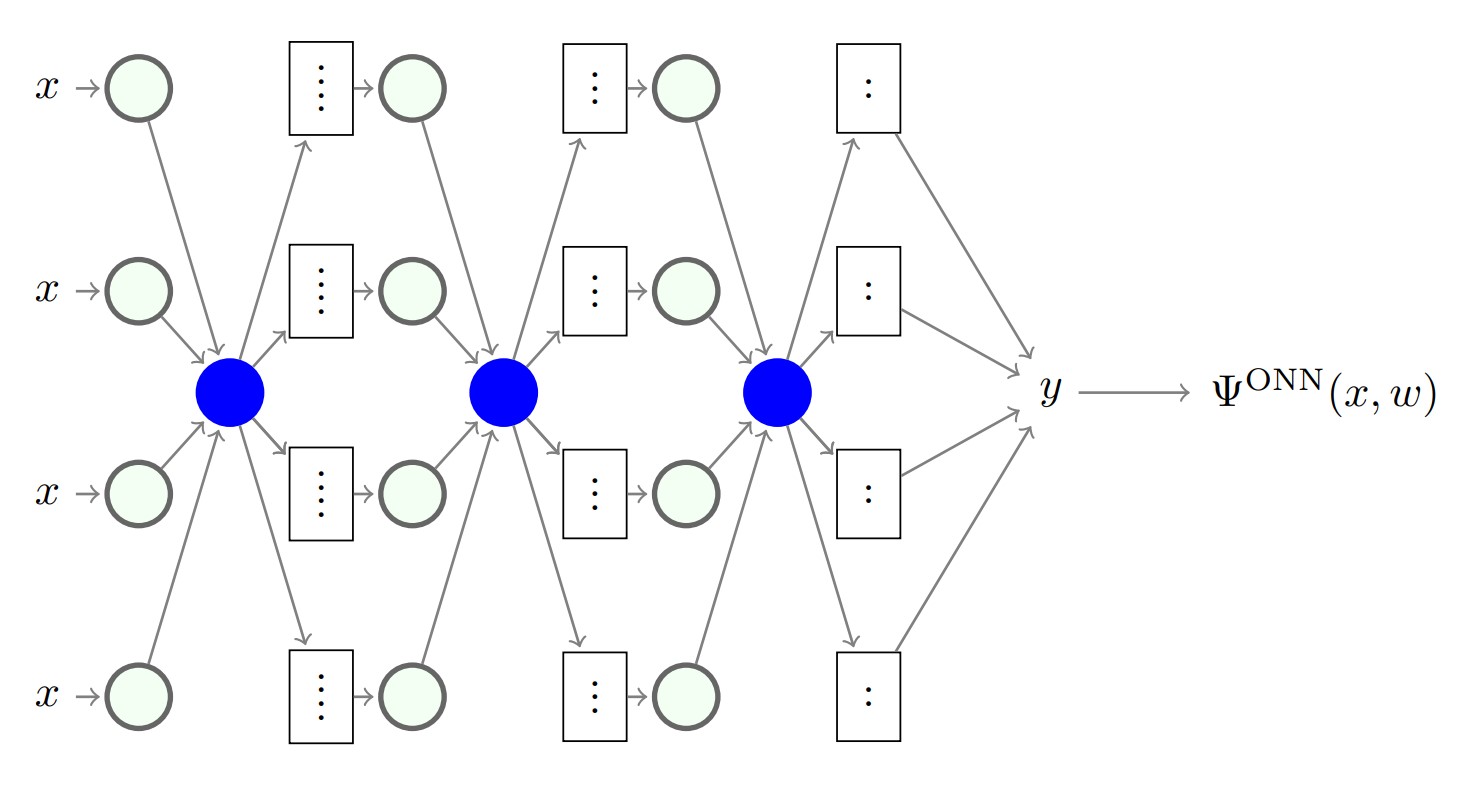}
	}
	\caption{
		(a)
		Base $4-3-2$ network, light circles indicate activations, boxes indicate post-activations.
		(b)
		Example for Design A with $2$ layers as input copies to each subsequent layer.
		The light circles indicate the linear operations/matrix-vector products.
		The results of the linear operation is averaged (single solid-blue circle) and fed through the activation function, producing the multiple version of the layers output (boxes).
		(c)
		Example of Design B.
	}
	\label{fig:Introduction__Schematic_depiction_of_the_two_noise_resistant_designs}
\end{figure}

Both designs---illustrated in \Cref{fig:Introduction__Schematic_depiction_of_the_two_noise_resistant_designs}---are built from a given \gls{NN} for which an optical implementation is desired.
Each design proposes a larger \gls{ONN} by taking parts of the original \gls{NN}, and duplicating and arranging them in a certain way.
If noise is absent, then this larger \gls{ONN} produces the same output as the original \gls{NN}; and, if noise is present, then this \gls{ONN} produces an output closer to the desired \gls{NN} than the direct implementation of the \gls{NN} as an \gls{ONN} without modifications  would give.

The first mechanism to construct a larger \gls{ONN} suppressing inherent noise of analog systems starts with a certain number of copies $N$ of the input data.
The copies are all processed independently by (in parallel arranged copies of) the layers.
Each copy of a layer takes in multiple input copies to produce the result of weighted addition, to which the activation mechanism is applied.
The copies that are transmitted to each layer (or set of parallel arrayed layers) are independent of each other.
The independent outputs function as inputs to the upcoming (copies of the) layers, and so on and so forth.

The idea of the second design is to use multiple copies of the input, on which the weighted addition is performed.
The noisy products of the weighted addition are averaged to a single number/light beam.
This average is then copied and the multiple copies are fed through the activation function, creating multiple noisy activations to be used as the next layer's input, and so on.

\subsection{Summary of results}
\label{sec:Introduction__Summary_of_results}

Using Design A, we are able to establish that \glspl{ONN} posses the same theoretical properties as \glspl{NN}.
Specifically, we can prove that \emph{any} \gls{NN} can be approximated arbitrarily well by an \gls{ONN} built using Design A (\Cref{thm:recursive_case}).
Similar considerations for \glspl{NN} with random edges can be found in \cite{manita2020universal}, but the noise model and proof method are different.
Here, we first bound the deviation of an \gls{ONN} and a noiseless \gls{NN}. To this bound Hoeffding’s inequality is then applied.

Establishing this theoretical guarantee, however, is done by increasing the number of components exponentially as the depth of the network increases.
The current proof shows that for an \gls{ONN} with Design A meant to approximate a \gls{NN} with $L$ layers arbitrarily well (and thus reduce the noise to negligible levels), a sufficient number of components is $\omega(K^{L(L+1)} L^L)$ for some constant $K > 0$. This is however not to say that such a large number is necessary: it is merely sufficient.

From a practical viewpoint, however, having to use as few components as possible would be more attractive.
We therefore also investigate Design B, in which the number of components increases only linearly with the depth of the network.
Because Design A already allows us to establish the approximation property of \glspl{ONN}, we limit our analysis of Design B to \emph{linear} \glspl{NN} for simplicity.
We specifically establish in \Cref{thm:Results__DesignB_recusive_structure_of_covariances} for any linear \gls{NN} the exact output distribution of an \gls{ONN} built using Design B.
Similar to the guarantee for Design A in \Cref{thm:recursive_case}, but more restrictively, this implies that any linear \gls{NN} can be approximated arbitrarily well by some \gls{ONN} built using Design B. Strictly speaking, Design B now has no guarantee of correctness for nonlinear \glspl{NN}, but this should practically not withhold us (especially when activations, for instance, are close to linear).

We conduct numerical experiments with Designs A and B by constructing to LeNet \glspl{ONN}.
The numerical results indicate that in practice, adding some components for noise negation is already sufficient to increase the accuracy of an \gls{ONN}; an exponential number does not appear not to be necessary (see \Crefrange{fig: LeNet_MSE_AvsB}{fig: LeNet_acc_AvsB}).

Finally, we want to remark that the high bandwidth of photonic circuits can be exploited to implement the designs as efficiently as possible.

\subsection{Outline of the paper}

We introduce the \gls{AWGN} model formally in \Cref{sec:Modeling_noise_in_ONNs}.
This model is the basis for the analysis of the proposed noise reduction schemes that are next discussed in \Cref{sec:Results__Design_A,sec:Results__Design_B}.
There, we specifically define Designs A and B, and each design is followed by a mathematical analysis.
The main results are \Cref{thm:recursive_case,thm:Results__DesignB_recusive_structure_of_covariances}.
\Cref{sec:Numerical_analysis} contain numerical simulations on LeNet \glspl{ONN} to which we apply Designs A and B.
\Cref{sec:Discussion} concludes; technical details are deferred to the Appendix.

\section{Model}
\label{sec:Modeling_noise_in_ONNs}
We consider general feed-forward \glspl{NN} implemented on analog optical devices.
Noise occurs due to various reasons in those optical devices.
Reasons include quantum noise in modulation, chip imperfections, and crosstalk
 \cite{tait2017neuromorphic, li2007channel, essiambre2010capacity, de2019noise, chakraborty2018toward}.

The noise profiles and levels of different devices differ, but we can, to good approximation, expect \gls{AWGN} to occur at three separate instances \cite{passalis2021training}: when modulating, when weighting, and when applying an activation function.
The thus proposed \gls{AWGN} model is formalized next in \Cref{sec:Feed_forward_nonlinear_ONNs}.

\subsection{Feed-forward nonlinear \texorpdfstring{\glspl{ONN}}{ONNs}}
\label{sec:Feed_forward_nonlinear_ONNs}

We assume that our aim is to implement a feed-forward nonlinear \gls{NN} with domain $\R^{d_0}$ and range $\R^{d_L}$, that can be represented by a parameterized function $\Psi^{\mathrm{NN}} : \R^{d_0} \times \R^n \rightarrow \R^{d_L}$ as follows.
For $\ell = 1, \ldots, L \in \naturalNumbersPlus$, $\Psi^{\mathrm{NN}}$ must be the composition of the functions
\begin{equation}
	\Psi^{\mathrm{NN}}_\ell : \R^{d_{\ell-1}}\rightarrow \R^{d_\ell}
	,
	\quad
	x
	\mapsto
	\sigma^{(\ell)} \bigl( W^{(\ell)}x+b^{(\ell)} \bigr).
	\label{eqn:NN_layers_as_functions}
\end{equation}
Here, $W^{(\ell)} \in \R^{d_\ell \times d_{\ell-1}}$ denotes the weight matrix in the $\ell$-th layer, $b^{(\ell)} \in \R^{d_\ell \times 1}$ the bias vector in the $\ell$-th layer, and $\sigma^{(\ell)}: \R^{d_\ell \times 1} \rightarrow \R^{d_\ell \times 1}$ the activation function in the $\ell$-th layer.
Specifically, the \gls{NN} satisfies
\begin{equation}
	\Psi^{\mathrm{NN}}(\cdot\, , w)
	=
	\Psi^{\mathrm{NN}}_L(\cdot \, , w^{(L)}) \circ \dots \circ \Psi^{\mathrm{NN}}_1(\cdot \, , w^{(1)})
	,
	\label{eqn:NN_as_a_composition_of_functions}
\end{equation}
where $w^{(\ell)} = (W^{(\ell)}, b^{(\ell)})$ represents the parameters in the $\ell$-th layer.
Note that we do not necessarily assume that the activation function is applied component-wise (it could be any high-dimensional function).
Such cases are simply contained within the model.

Suppose now that the \gls{NN} in \eqref{eqn:NN_as_a_composition_of_functions} is implemented as an \gls{ONN}, but \emph{without} amending its design.
\gls{AWGN} will then disrupt the output of each layer.
Specifically, for depths $L \in \naturalNumbersPlus$, the \gls{ONN} will be representable by a function $\Psi^{\mathrm{ONN}}$ that is the composition of the noisy functions
\begin{equation}
	\Psi_\ell^{\mathrm{ONN}} : \R^{d_{\ell-1}}\rightarrow \R^{d_\ell}
	,
	\quad
	x
	\mapsto
	\sigma^{(\ell)} \bigl( W^{(\ell)}x+b^{(\ell)}+ N_{\mathrm{w}}^{(\ell)} \bigr)
	+ N_{\mathrm{a}}^{(\ell)}
	\label{eqn:ONN_layers_as_functions}
\end{equation}
for $\ell = 1, \ldots, L \in \naturalNumbersPlus$.
Here,
\begin{equation}
	N_{\mathrm{w}}^{(\ell)}
	\eqcom{d}=
	\mathrm{Normal}(0, \Sigma_{\text{w}}^{(\ell)} )
	\textnormal{ and }
	N_{\mathrm{a}}^{(\ell)}
	\eqcom{d}=
	\mathrm{Normal}(0, \Sigma_{\text{act}}^{(\ell)} )
\end{equation}
denote multivariate normal distributions that describe the \gls{AWGN} within the \gls{ONN}.
In other words, the \gls{ONN} will satisfy
\begin{equation}
	\Psi^{\mathrm{ONN}}(\cdot\, , w)
	=
	\Psi_L^{\mathrm{ONN}}(\cdot \, , w^{(L)}) \circ \dots \circ \Psi_1^{\mathrm{ONN}}(\cdot \, , w^{(1)})
	\label{eqn:ONN_as_a_composition_of_functions}
\end{equation}
instead of \eqref{eqn:NN_as_a_composition_of_functions}.
Observe that \eqref{eqn:ONN_as_a_composition_of_functions} is a random \gls{NN}; its outcome is uncertain, but hopefully close to that of \eqref{eqn:NN_as_a_composition_of_functions}.

\subsection{Feed-forward linear \texorpdfstring{\glspl{ONN}}{ONNs}}
\label{sec:Feed_forward_linear_ONNs}

Let us briefly examine the special case of a feed-forward \emph{linear} \gls{ONN} in more detail.
That is, we now assume additionally that for $\ell = 1, \ldots, L$, there exist ${e}^{(\ell)} \in \mathbb{R}^{d_\ell}$ such that
$
	\sigma^{(\ell)}(y)
	=
	D^{(\ell)} y
$
where
$
	D^{(\ell)}
	=
	\mathrm{diag}(e^{(\ell)})
	.
$
In other words, each activation function $\sigma^{(\ell)}$ does element-wise multiplications by  constants.

If each activation function is linear, then the output distribution of each layer will remain multivariate normal distributed due to the so-called linear transformation theorem
\cite[Theorem 1.2.6]{muirhead2009aspects}.
The mean and covariance matrix of the underlying multivariate normal distribution will however be transformed in each layer.

Let us illustrate how the covariance matrix transforms by discussing the first layer in detail.
Each layer in \eqref{eqn:ONN_layers_as_functions} can be interpreted as a random function that takes the noisy vector $\mathbf{A}^{(\ell-1)}=(\mathbf{A}^{(\ell-1)}_1,\dots,\mathbf{A}^{(\ell-1)}_{d_{\ell-1}})$ say as input, and produces the even noisier vector $\mathbf{A}^{(\ell)} =(\mathbf{A}^{(\ell)}_1,\dots,\mathbf{A}^{(\ell)}_{d_\ell})$ say as output.
Specifically, the noisy input to the first layer is modeled by
\begin{equation}
	\mathbf{A}^{(0)} \mid x
	:\eqcom{d}=
	x + \mathcal{N}\left( 0, \Sigma_{\mathrm{m}} \right)
	\label{eqn:Noisy_input_to_the_first_layer}
\end{equation}
because of the modulation error within the first layer.
Here $\cdot \mid \cdot$ indicates a conditional random variable.
This input next experiences weighted addition and more noise is introduced: the noisy preactivation of the first layer satisfies
\begin{equation}
	\mathbf{U}^{(1)} \mid \mathbf{A}^{(0)}
	:\eqcom{d}=
	W^{(1)} \mathbf{A}^{(0)}
	+ b^{(1)}
	+ \mathcal{N}( 0, \Sigma_{\mathrm{w}}^{(1)} )
	.
	\label{eqn:Noisy_output_of_the_first_layer}
\end{equation}
Combining \eqref{eqn:Noisy_input_to_the_first_layer} and \eqref{eqn:Noisy_output_of_the_first_layer} with the linear transformation theorem for the multivariate normal distribution
as well as the fact that sums of independent multivariate normal random variables are again multivariate normally distributed \cite[Theorem 1.2.14]{muirhead2009aspects}, we find that
\begin{align}
	\mathbf{U}^{(1)} \mid x
	 &
	\eqcom{d}=
	W^{(1)} x
	+ b^{(1)}
	+ W^{(1)} \mathcal{N}\bigl( 0, \Sigma_{\mathrm{m}} \bigr)
	+ \mathcal{N}\bigl( 0, \Sigma_{\mathrm{w}}^{(1)} \bigr)
	\nonumber \\  &
	   \eqcom{d}=
	   W^{(1)} x
	   + b^{(1)}
	   + \mathcal{N}\bigl( 0, W^{(1)} \Sigma_{\mathrm{m}} ( W^{(1)} )^\intercal + \Sigma_{\mathrm{w}}^{(1)} \bigr)
	   .
\end{align}
After applying the linear activation function, we obtain
\begin{align}
	 &
	\mathbf{A}^{(1)} \mid x
	\eqcom{d}=
	\sigma_1 (\mathbf{U}^{(1)})
	+ \mathcal{N}( 0, \Sigma_{\mathrm{a}}^{(1)} )
	\mid x
	\nonumber \\  &
	   \eqcom{d}=
	   D^{(1)} (W^{(1)} x + b^{(1)} )
	\nonumber \\  &
	   \phantom{\eqcom{d}=}
	   +
	   \mathcal{N}
	   \Bigl(
	   0
	   ,
	   \Sigma_{\mathrm{a}}^{(1)}
	   +
	   D^{(1)}
	   \bigl(
		   W^{(1)} \Sigma_{\mathrm{m}} (W^{(1)})^\intercal
		   +
		   \Sigma_{\mathrm{w}}^{(1)}
		   \bigr)
		   (D^{(1)})^\intercal
	   \Bigr)
	\nonumber \\  &
	   =
	   \mathcal{N}
	   (
	   \Psi^{\mathrm{NN}}_1( x, w ),
	   \Sigma^{(1)}_{\mathrm{ONN}}
	   )
	   \label{eqn:Output_of_the_first_layer_in_an_ONN}
\end{align}
say.
Observe that the unperturbed network's output remains intact, and is accompanied by a centered normal distribution with an increasingly involved covariance matrix:
\begin{align}
	 &
	\Sigma_{\mathrm{ONN}}^{(1)}
	\nonumber \\  &
	   =
	   D^{(1)} \bigl( W^{(1)} \Sigma_{\mathrm{m}} (W^{(1)})^\intercal + \Sigma_{\mathrm{w}}^{(1)} \bigr) (D^{(1)})^\intercal
	   + \Sigma_{\mathrm{a}}^{(1)}
	\nonumber \\  &
	   =
	   D^{(1)} W^{(1)} \Sigma_{\mathrm{m}} (W^{(1)})^\intercal (D^{(1)})^\intercal + D^{(1)}\Sigma_{\mathrm{w}}^{(1)} (D^{(1)})^\intercal
	\nonumber \\  &
	   \qquad+
	   \Sigma_{\mathrm{a}}^{(1)}
	   .
	   \label{eqn:Covariance_matrix_in_a_linear_one_layer_NN}
\end{align}
Observe furthermore that the covariance matrix in \eqref{eqn:Covariance_matrix_in_a_linear_one_layer_NN} is independent of the bias $b^{(1)}$.

The calculations in \crefrange{eqn:Noisy_input_to_the_first_layer}{eqn:Covariance_matrix_in_a_linear_one_layer_NN} can readily be extended into a recursive proof that establishes the covariance matrix of the entire linear \gls{ONN}.
Specifically, for $\ell = 1, \ldots, L$, define the maps
\begin{align}
	T^{(\ell)}( \Sigma )
	 &
	:=
	D^{(\ell)} W^{(\ell)} \Sigma (W^{(\ell)})^\intercal (D^{(\ell)})^\intercal
	\nonumber \\  &
	   \phantom{:=}
	   +
	   D^{(\ell)} \Sigma_{\mathrm{w}}^{(\ell)} (D^{(\ell)})^\intercal
	   +
	   \Sigma_a^{(\ell)}
	   .
	   \label{eq:Model__definition_iteration_map}
\end{align}
We then have the following:

\begin{proposition}[Distribution of linear \glspl{ONN}]
	\label{prop:Model__recusive_structure_of_covariances}

	Assume that there exist vectors $e^{(\ell)}\in \R^{d_\ell}$ such that $\sigma^{(\ell)}(y) \allowbreak = \allowbreak \textrm{diag}(e^{(\ell)}) y$.
	The feed-forward linear \gls{ONN} in \eqref{eqn:ONN_as_a_composition_of_functions} then satisfies
	\begin{equation}
		\Psi^{\mathrm{ONN}}(\cdot, w)
		\eqcom{d}=
		\mathcal{N}
		\bigl(
		\Psi^{\mathrm{NN}}(\cdot, w)
		,
		\Sigma_{\mathrm{ONN}}^{(L)}
		\bigr)
		,
	\end{equation}
	where for $\ell = L, L-1, \ldots, 1$,
	\begin{equation}
		\Sigma_{\mathrm{ONN}}^{(\ell)}
		=
		T^{(\ell)}( \Sigma_{\mathrm{ONN}}^{(\ell-1)} )
		;
		\quad
		\textnormal{and}
		\quad
		\Sigma_{\mathrm{ONN}}^{(0)}
		= \Sigma_{\mathrm{m}}
		.
	\end{equation}
\end{proposition}

In linear \glspl{ONN} with symmetric noise (that is, the \gls{AWGN} of  each layer's noise sources has the same covariance matrix), \Cref{prop:Model__recusive_structure_of_covariances}'s recursion simplifies.
Introduce $P^{(\ell)} := \prod_{i=\ell+1}^{L}D^{(i)} W^{(i)}$ for notational convenience.
The following is proved in \Cref{sec:Limit_for_symmetric_linear_ONNs}:

\begin{corollary}[Symmetric noise case]
	\label{cor:Limit_for_symmetric_linear_ONNs}

	Within the setting of \Cref{prop:Model__recusive_structure_of_covariances},
	assume additionally that for all $\ell \in \naturalNumbersPlus$,
	$\Sigma_{\mathrm{a}}^{(\ell)} = \Sigma_{\mathrm{a}}$
	and
	$\Sigma_{\mathrm{w}}^{(\ell)} = \Sigma_{\mathrm{w}}$.
	Then,
	\begin{align}
		\Sigma_{\mathrm{ONN}}^{(L)}
		 &
		=
		P^{(0)}
		\Sigma_\mathrm{m}
		( P^{(0)} )^\intercal
		+
		\sum_{\ell=1}^{L} P^{(\ell)} \Sigma_\mathrm{a} (P^{(\ell)})^\intercal
		\nonumber \\  &
		   \phantom{=}
		   +
		   \sum_{\ell=1}^{L} P^{(\ell)} D^{(\ell)}\Sigma_\mathrm{w}(D^{(\ell)})^\intercal \bigl(P^{(\ell)}\bigr)^\intercal
		   .
	\end{align}
	If moreover for all $\ell \in \naturalNumbersPlus$,
	$W^{(\ell)} = W$,
	$D^{(\ell)} = D$,
	and
	$\pnorm{ D }{\mathrm{F}} \pnorm{W}{\mathrm{F}} < 1$,
	then
	\begin{align}
		\lim_{L \to \infty}
		\Sigma^{(L)}_{\mathrm{ONN}}
		=
		\sum_{n=0}^\infty (DW)^n \left(D \Sigma_\mathrm{w} D^\intercal + \Sigma_\mathrm{a}\right) \left((DW)^n\right)^\intercal.
	\end{align}
\end{corollary}

\Cref{prop:Model__recusive_structure_of_covariances} and \Cref{cor:Limit_for_symmetric_linear_ONNs} describe the output distribution of linear \glspl{ONN} completely.

\subsection{Discussion}

One way to think of the \gls{AWGN} model in \Cref{sec:Feed_forward_nonlinear_ONNs} is to take a step back from the microscopic analysis of individual devices, and consider an \gls{ONN} as a series of black box devices (recall also \Cref{fig: possible_fig_for_modelling_section}).
Each black box device performs their designated task and acts as communication channels with \gls{AWGN}.
This way of modeling in order to analyze the impact of noise can also be seen in \cite{passalis2021training}; and other papers modeling optical channels include \cite{li2007channel, essiambre2010capacity}.
Further papers considering noise in optical systems with similar noise assumptions are \cite{semenova2022understanding, semenova2019fundamental}, where furthermore multiplicative noise is considered when an amplifier is present within the circuit \cite{semenova2022understanding}. Qualitatively the results for Design A also apply for multiplicative noise, the scaling however may differ.

\paragraph{Limitations of the model}
We note firstly that modeling the noise in \glspl{ONN} as \gls{AWGN} is warranted only in an operating regime with many photons, and is thus unlikely to be a good model for \glspl{ONN} that operate in a regime with just a few photons.

Secondly, due to physical device features and operation conditions, weights, activations, and outputs can only be realized in \glspl{ONN} if their values lie in certain ranges.
Such constraints are no part of the model in \Cref{sec:Modeling_noise_in_ONNs}.
Fortunately, however, the implied range restrictions are usually not a problem in practice.
For example, activation functions like sigmoid and $\tanh$ map into $[0,1]$ and $[-1,1]$, respectively.
Additional regularization rules like weight decay also move the entries of weight matrices in \glspl{NN} towards smaller values.
In case physical constraints were met one can increase the weight decay parameter to further penalize large weights during training, leading to smaller weights so that the \gls{ONN} is again applicable.

\section{Results---Design A}
\label{sec:Results__Design_A}

\subsection{Reducing the noise in feed-forward \texorpdfstring{\glspl{ONN}}{ONNs} (Design A)}
\label{sec:Results__Definition_of_Design_A}

Recall that an example of Design A is presented in \Cref{fig:Introduction__Schematic_depiction_of_the_two_noise_resistant_designs}(b).
\Cref{alg:Results__noise_reduction_tree_NN} constructs this tree-like network, given the desired number of copies $n_0, \ldots, n_L$ per layer.

\begin{algorithm}
  \caption{Algorithm to construct a noise reducing network}
  \label{alg:Results__noise_reduction_tree_NN}

  \begin{algorithmic}
    \REQUIRE Input $\mathbf{n}=(n_\ell)_{\ell=0,\dots,L}$
    \REQUIRE $\prod_{\ell=0}^Ln_i$ copies of input $x^{(0)}$, named $^1x^{(0)},\dots,^{\big(\prod_{\ell=0}^Ln_i\big)}x^{(0)}$
    \FOR{$\ell=0,\dots , L-1$}
	    \FOR{$\alpha = 1,\dots , \prod_{i=\ell}^{L-1}n_i$}
		    \STATE $^{\alpha}\xi^{(\ell)} \gets W^{(\ell+1)} \,^{\alpha}x^{(\ell)} + b^{(\ell+1)} +\mathrm{Normal}(0,\Sigma_\text{w})$
	    \ENDFOR
	    \FOR{$\alpha=0,\dots,\big(\prod_{i=\ell}^{L-1}n_i\big)-1$}
		    \STATE $^{\alpha}y^{(\ell)} \xleftarrow{\mathrm{averaging}} n_{\ell}^{-1} \,\left(^{\alpha \cdot n_{\ell} + 1}\xi^{(\ell)} + \dots + ^{\alpha \cdot n_{\ell} + n_{\ell}}\xi^{(\ell)} \right)$
		    \STATE $^{\alpha}x^{(\ell+1)} \gets \sigma^{(\ell)}( ^{\alpha} y^{(\ell)})+\mathrm{Normal}(0,\Sigma_\text{a})$
	    \ENDFOR
    \ENDFOR
    \RETURN $^1x^{(L)}$
  \end{algorithmic}
\end{algorithm}

Observe that in Design A, the number of copies utilized in each layer, the $n_\ell$, are fixed.
There is however only a single copy in the last layer.
Its output is the unique output of the \gls{ONN}.
Each other layer receives multiple independent inputs.
With each of the independent copies weighted addition is performed, and the results are averaged to produce the layer's single output.
Having independent incoming copies is achieved by having multiple independent branches of the prior partial networks incoming into a given layer.
This means that the single layer $L$ receives $n_{L-1}$ independent inputs of $n_{L-1}$ independent layers $L-1$.
Each of the $n_{L-1}$ copies of layer $L-1$ receives $n_{L-2}$ inputs from independent copies of layer $L-2$.
Generally, let $n_{\ell-1}$ be the number of copies of layer $\ell-1$ that act as inputs to layer $\ell$.

Observe that all copies are created upfront.
That means there are $\prod_{\ell=0}^{L-1}n_\ell$ copies of the data.
By \Cref{alg:Results__noise_reduction_tree_NN}, $\prod_{\ell=1}^{L-1}n_\ell$ copies of the first layer are arrayed in parallel to each other, and each of them processes $n_0$ copies of the data.
The outputs of the $\prod_{\ell=1}^{L-1}n_\ell$ arrayed copies of the first layer are the input to the $\prod_{\ell=2}^{L-1}n_\ell$ arrayed copies of the second layer, and so on.

Notice that noise stemming from applying the activation function is subject to a linear transformation in the next layer.
The activation function noise can therefore be considered as weight-noise by inserting an identity layer with $\sigma=\mathrm{id}$, $W=I$ and $b=0$.

We want to verify that a Design A \gls{ONN} yields outputs that are with high probability close to the original noiseless \gls{NN}.
Let $\tilde{\Psi}^{\mathrm{ONN}}(x, w)$ the Design A \gls{ONN} and then let
\begin{align}
	\mathbbm{P}
	\Bigl[
	\sup_{x\in \R^d } \big\| \Psi^{\mathrm{NN}}(x, w) -\tilde{\Psi}^{\mathrm{ONN}}(x, w)\big\|_2
	<
	D_L
	\Bigr]
	>
	1-C_L,
	\label{eq:DesignA_desired_prop}
\end{align}
be the desired property.
The main result of this section is the following:
 
 \begin{theorem}\label{thm:recursive_case}
 	For any $C_L \in (0,1)$, any $D_L\in (0,\infty)$, and any nonlinear \gls{NN} $\Psi^{\mathrm{NN}}$, with Lipschitz-continuous activations functions with Lipschitz-constants $a^{(i)}$ and weight-matrices $W^{(i)}$,
 	\Cref{alg:Results__noise_reduction_tree_NN} is able to construct an \gls{ONN} $\tilde{\Psi}^{\mathrm{ONN}}$ that satisfies \eqref{eq:DesignA_desired_prop}.
 	
 	Let the covariance matrices of the occurring \gls{AWGN} be diagonal matrices and let each of the values of the covariance matrices be upper bounded by $\sigma^2\geq 0$.
 	For any set of $(\kappa_i)_{i=1,\dots,L}$, $(\delta_i)_{i=1,\dots,L}$ such that $\prod (1-\kappa_\ell) > 1-C_L$ and
 	$\sum \delta_\ell \leq D_L$,
 	a sufficient number of copies to construct an \gls{ONN} $\tilde{\Psi}^{\mathrm{ONN}}$ that satisfies \eqref{eq:DesignA_desired_prop}
 	is given by
 	\begin{align}
 		n_L
 		&=
 		1
 		\\
 		n_{\ell}
 		&\geq
 		\frac{
 			\sigma^2
 			\Bigl(\prod_{i=\ell+1}^{L}a^{(i)}
 			\prod_{i=\ell+2}^{L}\|W^{(i)}\|_\mathrm{op}\Bigr)^2
 		}{
 			\delta_{\ell+1}^2
 		}
 		\\
 		&
 		\phantom{\geq}
 		\times
 		\Biggl(
 		\sqrt{2}
 		\frac{
 			\Gamma ((d_{\ell+1}+1)/2)
 		}{
 			\Gamma (d_{\ell+1}/2)
 		}
 		\\
 		&
 		\phantom{\geq\times}
 		+
 		\sqrt{
 			\frac{C^2}{c}
 			\frac{4\sqrt[d_{\ell+1}]{4}}{2\sqrt[d_{\ell+1}]{4}-2}
 			(-\ln\bigl(\kappa_{\ell+1}/2\bigr))
 			\frac{1}{
 				\prod_{i=\ell +1}^L n_i}
 		}
 		\Biggr)^2,
 		\\
 		&
 		\qquad\qquad\qquad\qquad\qquad\qquad\qquad
 		\ell=L-1,\dots,0.
 	\end{align}
 Here $\Gamma$ is the gamma function and $C,c>0$ are absolute constants.
 \end{theorem}

This result is proven in \Cref{sec:Design_A__Analysis_of_Design_A}.
A consideration on the asymptotic total amount of copies in deep \glspl{ONN} is relegated to \Cref{appendix: Design A}.

\subsection{Idea behind Design A}
\label{sec:Design_A__Motivation_of_Design_A}

Having the law of large numbers in mind it seems reasonable that the average of multiple experiments would help in achieving a more precise output in the presence of noise.
However, it would typically not be correct to just input $n$ identical, deterministic copies of $x$ into $n$ independent \glspl{ONN}---thus producing $n$ noisy realizations $\Psi^{\mathrm{ONN},1}(x, w), \ldots,\Psi^{\mathrm{ONN},n}(x, w)$ say---and then calculate their average in the hope to recover $\Psi^{\mathrm{NN}}(x,w)$.
This is because while by the law of large numbers it is true that
\begin{equation}
  \lim_{n\rightarrow\infty} \frac{1}{n}\sum_{i=1}^n \Psi^{\mathrm{ONN},i}(x, w)
  =
  \mathbbm{E}\left[ \Psi^{\mathrm{ONN}}(x, w) \right]
  ,
\end{equation}
it is not necessarily true that the expectation
$
  \mathbbm{E}\left[ \Psi^{\mathrm{ONN}}(x, w) \right]
$
equals
$
  \Psi^{\mathrm{NN}}(x, w)
$.
The reason is that activation functions in \glspl{NN} are typically nonlinear.

We can circumvent the issue by modifying the approach and instead exploit the law of large numbers layer-wise.
Recall that in the noiseless \gls{NN}, layer $\ell$ maps a fixed input
\begin{align}
  x
  \mapsto
  \sigma^{(\ell)}(W^{(\ell)} x+b^{(\ell)} )
  ,
  \label{unperturbed output layer L}
\end{align}
and that the same layer in the \gls{ONN} maps the same fixed input
\begin{align}
  x
  \mapsto
  \sigma^{(\ell)}(W^{(\ell)} x+b^{(\ell)} + {N}_{\mathrm{w}}^{(\ell)} )
\end{align}
instead.
If we let $(N^{(i)})_{i\in \{1,\dots,n \}}$ be independent realizations of the distribution of ${N}_{\mathrm{w}}^{(\ell)}$ (which has mean zero), we can expect by the law of large numbers that for sufficiently large $n$, the realized quantities
\begin{equation}
  \frac{1}{n} \bigl(\sum_{i=1}^n W^{(\ell)} x+b^{(\ell)} + N^{(i)} \bigr)\quad
  \mathrm{ and } \quad W^{(\ell)} x+b^{(\ell)}
\end{equation}
are close to each other.
If $\sigma^{(\ell)}$ is moreover sufficiently regular, then we may expect that the realized quantity
\begin{align}
  \sigma^{(\ell)}\Bigl(\frac{1}{n} \Bigl(\sum_{i=1}^n W^{(\ell)} x+b^{(\ell)} + N^{(i)} \Bigr)\Bigr)
  \label{corrected perturbed output layer L}
\end{align}
is close to \eqref{unperturbed output layer L} for sufficiently large $n$, i.e., close to the unperturbed output of the original layer.

The implementation in \eqref{corrected perturbed output layer L} can be realized by using $n$ times as many nodes in the hidden layer; thus to essentially create $n$ copies of the original hidden layer.
These independent copies are then averaged.
Furthermore, one can allow for different inputs $(x^i)_{i\in \{1,\dots , n \}}$, assuming some statistical properties of their distribution.
This will be formalized next in the proof of \Cref{thm:recursive_case} in \Cref{sec:Design_A__Analysis_of_Design_A}.

\subsection{Proof of \texorpdfstring{\Cref{thm:recursive_case}}{Theorem 1}}
\label{sec:Design_A__Analysis_of_Design_A}

For the proof we will first upper bound the deviation between an \gls{ONN} constructed with Design A and the noiseless \gls{NN} (\Cref{sec:__Design_A_ONN_NN_bound}) and then we find a probabilistic bound on the deviations bound (\Cref{sec:__Design_A_bound_for_deviations}).

\subsubsection{An upper-bound for the \texorpdfstring{\gls{ONN}}{ONN} --- \texorpdfstring{\gls{NN}}{NN} deviation}\label{sec:__Design_A_ONN_NN_bound}

The output of the Design A network is
\begin{align}
	\tilde{x}
	=
	\sigma^{(L)}
	\Bigl(
	\frac{1}{n_{L-1}}
	\sum_{i=1}^{n_{L-1}}
	\bigl(
	W^{(L)} \tilde{x}^{i} + b^{(L)}
	+ N^{(i)}
	\bigr)
	\Bigr),
	\label{eq:appendix_ONN_calc}
\end{align}
where each $\tilde{x}^{i}$ is recursively calculated as
\begin{align}
	\tilde{x}^{i}
	=
	\sigma^{(L-1)}
	\Bigl(
	\frac{1}{n_{{L-2}}}
	\sum_{j_{i}=1}^{n_{L-2}}
	\Bigl(
	W^{(L-1)}
	\tilde{x}^{j_i} + b^{({L-1})}
	+ N^{(j_i)}
	\Bigr)
	\Bigr),
\end{align}
the $\tilde{x}^{j_i}$ are calculated as
\begin{align}
	\tilde{x}^{j_i}
	=
	\sigma^{(L-2)}
	\Bigl(
	\frac{1}{n_{{L-3}}}
	\sum_{k_{j_i}=1}^{n_{L-3}}
	\Bigl(
	W^{(L-2)}
	\tilde{x}^{k_{j_i}} + b^{({L-2})}
	+ N^{(k_{j_i})}
	\Bigr)
	\Bigr),
\end{align}
and so on and so forth.
The difference in $L_2$-norm of \eqref{eq:appendix_ONN_calc} and the noiseless \gls{NN}
\begin{align}
	\sigma^{(L)}
	\bigl(
	W^{(L)}
	\sigma^{(L-1)}
	\bigl(
	W^{(L-1)}
	\bigl(
	\dots
	\bigr)
	+b^{(L-1)}
	\bigr)
	+b^{(L)}
	\bigr)
\end{align}
can iteratively be bounded by using the Lipschitz property of the activation functions, triangle inequality, and submultiplicativity of the norms.

We start the iteration by bounding
\begin{align}
&
	\phantom{\leq}
	\Bigl\|
	\sigma^{(L)}
	\Bigl(
	\frac{1}{n_{L-1}}
	\sum_{i=1}^{n_{L-1}}
	\Bigl(W^{(L)} \tilde{x}^{i} + b^{(L)}
	+ N^{(i)}\Bigr)
	\Bigr)
\\
&
	\phantom{\leq}
	-
	\sigma^{(L)}
	\bigl(
	W^{(L)}
	\sigma^{(L-1)}
	\bigl(
	W^{(L-1)}
	\bigl(
	\dots
	\bigr)
	+b^{(L-1)}
	\bigr)
	+b^{(L)}
	\bigr)
	\Bigr\|_2
\\
&
	\leq
	a^{(L)}
	\Bigl\|
	\frac{1}{n_{L-1}}
	\sum_{i=1}^{n_{L-1}}
\\
&
	\phantom{\leq}
	\Bigl
	(W^{(L)}
	\Bigl(
	\tilde{x}^{i}-
	\sigma^{(L-1)}
	\bigl(
	W^{(L-1)}
	\bigl(
	\dots
	\bigr)
	+b^{(L-1)}
	\bigr)
	\Bigr)
	+ N^{(i)}
	\Bigr)
	\Bigr\|_2
\\
&
	\leq
	\frac{
		a^{(L)}
		\|W^{(L)}\|_\mathrm{op}
	}{
		n_{L-1}
	}
\\
&
	\phantom{leq}
	\times
	\Bigl\|
	\sum_{i=1}^{n_{L-1}}
	\Bigl(
	\tilde{x}^{i}-
	\sigma^{(L-1)}
	\bigl(
	W^{(L-1)}
	\bigl(
	\dots
	\bigr)
	+b^{(L-1)}
	\bigr)
	\Bigr)
	\Bigr\|_2
\\
&
	\phantom{\leq}
	+
	a^{(L)}
	\Bigl\|
	\frac{1}{n_{L-1}}
	\sum_{i=1}^{n_{L-1}}
	N^{(i)}
	\Bigr\|_2.
\end{align}

In the next iteration step the term
\begin{align}
	\Bigl\|
	\sum_{i=1}^{n_{L-1}}
	\Bigl(
	\tilde{x}^{i}-
	\sigma^{(L-1)}
	\bigl(
	W^{(L-1)}
	\bigl(
	\dots
	\bigr)
	+b^{(L-1)}
	\bigr)
	\Bigr)
	\Bigr\|_2
\end{align}
is further bounded by first using the triangle inequality and thereafter bounding in the same way as we did in the first layer:
\begin{align}
&
	\Bigl\|
	\sum_{i=1}^{n_{L-1}}
	\Bigl(
	\sigma^{(L-1)}
	\Bigl(
	\frac{1}{n_{{L-2}}}
	\sum_{j_{i}=1}^{n_{L-2}}
	\Bigl(
	W^{(L-1)}
	\tilde{x}^{j_i} + b^{({L-1})}
	+ N^{(j_i)}
	\Bigr)
	\Bigr)
\\
&
	-
	\sigma^{(L-1)}
	\bigl(
	W^{(L-1)}
	\bigl(
	\sigma^{(L-2)}
	\bigl(
	W^{(L-2)}
	\bigl(
	\dots
	\bigr)
	+b^{(L-2)}
	\bigr)
	\bigr)
\\
&
	\phantom{-\sigma^{(L-1)}\bigl(}
	+b^{(L-1)}
	\bigr)
	\Bigr)
	\Bigr\|_2
\\
&
	\leq
	\frac{
		a^{(L-1)}
		\|W^{(L-1)}\|_\mathrm{op}
	}{
		n_{L-2}
	}
\\
&
	\phantom{\leq}
	\times
	\sum_{i=1}^{n_{L-1}}
	\Bigl\|
	\sum_{j_i=1}^{n_{L-2}}
	\Bigl(
	\tilde{x}^{j_i}-
	\sigma^{(L-2)}
	\bigl(
	W^{(L-2)}
	\bigl(
	\dots
	\bigr)
	+b^{(L-2)}
	\bigr)
	\Bigr)
	\Bigr\|_2
\\
&
	\phantom{\leq}
	+
	a^{(L-1)}
	\sum_{i=1}^{n_{L-1}}
	\Bigl\|
	\frac{1}{n_{L-2}}
	\sum_{j_i=1}^{n_{L-2}}
	N^{(j_i)}
	\Bigr\|_2.
\end{align}
Here,
\begin{align}
	\sum_{i=1}^{n_{L-1}}
	\Bigl\|
	\sum_{j_i=1}^{n_{L-2}}
	\Bigl(
	\tilde{x}^{j_i}-
	\sigma^{(L-2)}
	\bigl(
	W^{(L-2)}
	\bigl(
	\dots
	\bigr)
	+b^{(L-2)}
	\bigr)
	\Bigr)
	\Bigr\|_2
\end{align}
may again be bounded in the same fashion. This leads to the following recursive argument.

Let $\mathcal{F}^{(\ell)}$ be the sum of the differences between---loosely speaking---the ends of the remaining Design A ``subtrees'' and noiseless \glspl{NN} ``subtrees'' at layer $\ell$. More specifically, let
\begin{align}
	\mathcal{F}^{(L)}
&
	:=
	\Big\|
	\tilde{x}
	-
	\sigma^{(L)}
	\bigl(
	W^{(L)}
	\bigl(
	\dots
	\bigr)
	+b^{(L)}
	\bigr)
	\Big\|;
\\
	\mathcal{F}^{(\ell)}
&
	:=
	\sum_{i_{L}=1}^{n_{L-1}}\sum_{i_{{L-1}_L}=1}^{n_{L-2}}
	\dots
	\sum_{i_{{{\ell+2}_{\dots_{{L-1}_L}}}}=1}^{n_{\ell+1}}
	\Big\|
	\sum_{i_{{\ell+1}_{{\ell+2}_{\dots_{{L-1}_L}}}}=1}^{n_{\ell}}
\\
&
	\phantom{=}
	\qquad
	\Bigl(
	\tilde{x}^{i_{{\ell+1}_{{\ell+2}_{\dots_{{L-1}_L}}}}}
	-
	\sigma^{(\ell)}
	\bigl(
	W^{(\ell)}
	\bigl(
	\dots
	\bigr)
	+b^{(\ell)}
	\bigr)
	\Bigr)
	\Big\|_2,
\\
&
	\qquad\qquad\qquad\qquad\qquad\qquad\qquad
	\forall \ell = 1,\dots,L-1,
\end{align}
the special case of $\mathcal{F}^{(0)}$ will be considered in detail later.
For simplicity, we join the sums outside the norm into one. Notice that because $n_L=1$, we have $\prod_{k=\ell+1}^{L-1} n_k=\prod_{k=\ell+1}^{L} n_k$, and we can write
\begin{align}
	\mathcal{F}^{(\ell)}
	=
	\sum_{i=1}^{\prod_{k=\ell+1}^{L} n_k}
	\Big\|
	\sum_{j_i=1}^{n_\ell}
	\Bigl(
	\tilde{x}^{j_i}
	-
	\sigma^{(\ell)}
	\bigl(
	W^{(\ell)}
	\bigl(
	\dots
	\bigr)
	+b^{(\ell)}
	\bigr)
	\Bigr)
	\Big\|_2,
\end{align}
where specifically
\begin{align}
	\tilde{x}^{j_i} =
	\sigma^{(\ell)}
	\Bigl(
	\frac{1}{n_{{\ell-1}}}
	\sum_{k_{j_{i}}=1}^{n_{\ell-1}}
	\Bigl(
	W^{(\ell)}
	\tilde{x}^{k_{j_i}} + b^{({\ell})}
	+ N^{(k_{j_i})}
	\Bigr)
	\Bigr),
\end{align}
and the $j_i$ and $k_{j_i}$ are nothing more than relabelings.

Bounding $\mathcal{F}^{(\ell)}$ using the triangle inequality, Lipschitz-property, and submultiplicativity yields
\begin{align}
	\mathcal{F}^{(\ell)}
&
	\leq
	a^{(\ell)}
	\sum_{i=1}^{\prod_{k=\ell+1}^{L} n_k}
	\sum_{j_i=1}^{n_\ell}
	\Bigl\|
	\frac{1}{n_{\ell-1}}
	\sum_{k_{j_i}=1}^{n_{\ell-1}} \Bigl(N^{(k_{j_i})}
\\
&
	\phantom{\leq}
	+W^{(\ell)}
	\Bigl(
	\tilde{x}^{k_{j_i}}-
	\sigma^{(\ell-1)}
	\bigl(
	W^{(\ell-1)}
	\bigl(
	\dots
	\bigr)
	+b^{(\ell-1)}
	\bigr)
	\Bigr)\Bigr)
	\Bigr\|_2
\\
&
	\leq
	\frac{
		a^{(\ell)}
		\|W^{(\ell)}\|_\mathrm{op}
	}{
		n_{\ell-1}
	}
	\mathcal{F}^{(\ell-1)}
\\
&
	\phantom{\leq}
	+
	a^{(\ell)}
	\sum_{i=1}^{\prod_{k=\ell+1}^{L} n_k}
	\sum_{j_i=1}^{n_\ell}
	\Bigl\|
	\frac{1}{n_{\ell-1}}
	\sum_{k_{j_i}=1}^{n_{\ell-1}}
	N^{(k_{j_i})}
	\Bigr\|_2.
	\label{eq:iteration}
\end{align}
We thus found a recursive formula for the bound.

The recursion ends at $\mathcal{F}^{(0)}$. The noiseless \gls{NN} receives $x$ as input, while the \gls{ONN} receives modulated input $x+N^{(j_i)}$, where $N^{(j_i)}$ is the modulation noise, i.e., \gls{AWGN}. Therefore,
\begin{align}
	\mathcal{F}^{(0)}
&
	=
	\sum_{i=1}^{\prod_{k=1}^{L} n_k}
	\Big\|
	\sum_{j_i=1}^{n_0}
	((x+N^{(j_i)})
	-
	x
	)
	\Big\|_2
\\
&
	=
	\sum_{i=1}^{\prod_{k=1}^{L} n_k}
	\Big\|
	\sum_{j_i=1}^{n_0}
	N^{(j_i)}
	\Big\|_2.
	\label{eq: x dependence disappears}
\end{align}
Observe that the $x$-dependence disappeared.

Readily iterating \eqref{eq:iteration} leads to the bound
\begin{align}
&
	\phantom{\leq}
	\mathcal{F}^{(L)}
\\
&
	\leq
	\sum_{\ell=L,L-1,\dots,1} \prod_{i=\ell}^{L}a^{(i)} \prod_{i=\ell+1}^{L}\|W^{(i)}\|_\mathrm{op}
\\
&
	\phantom{=\sum_{\ell=L,L-1,\dots,1}}
	\times
	\frac{1}{\prod_{k=\ell}^{L}n_{k}}\frac{1}{n_{\ell-1}} \sum_{i=1}^{\prod_{k=\ell}^{L}n_{k}}
	\Bigl\|
	\sum_{j_i=1}^{n_{\ell-1}}N^{(j_i)}
	\Bigr\|_2.
\end{align}
Therefore, if all the $L_2$-norms of the sums of the Gaussians are small at the same time, the network is close to the noiseless \gls{NN}. Let
\begin{align}
&
	\mathcal{S}_\ell := \prod_{i=\ell}^{L}a^{(i)}
	\prod_{i=\ell+1}^{L}\|W^{(i)}\|_\mathrm{op}
\\
&
	\phantom{=\sum_{\ell=L,L-1,\dots,1}}
	\times
	\frac{1}{\prod_{k=\ell}^{L}n_{k}}\frac{1}{n_{\ell-1}} \sum_{i=1}^{\prod_{k=\ell}^{L}n_{k}} 
	\Bigl\|
	\sum_{j_i=1}^{n_{\ell-1}}N^{(j_i)}
	\Bigr\|_2.
\end{align}
If for all $\ell$
\begin{align}
	\mathbb{P}\bigl[ \mathcal{S}_\ell \leq \delta_\ell \bigr] > 1-\kappa_\ell,\label{eq: single layer condition}
\end{align}
and moreover $\sum \delta_\ell \leq D_L$ as well as $\prod (1-\kappa_\ell) > 1-C_L$,
then \eqref{eq:DesignA_desired_prop} holds.
This can be seen by bounding
\begin{align}
&
	\phantom{\geq}
	\mathbbm{P}
	\Bigl[
	\sup_{x\in \R^d } \big\| \Psi^{\mathrm{NN}}(x, w) -\tilde{\Psi}^{\mathrm{ONN}}(x, w)\big\|_2
	<
	D_L
	\Bigr]
\\
&
	\geq
	\mathbbm{P}
	\Bigl[
	\sum_\ell \mathcal{S}_\ell
	< D_L
	\Bigr]
\geq
	\mathbbm{P}
	\Bigl[
	\bigcap_\ell
	\bigl\{
	\mathcal{S}_\ell
	< \delta_\ell
	\bigr\}
	\Bigr]
\\
&
	=
	\prod_\ell
	\mathbbm{P}
	\Bigl[
	\mathcal{S}_\ell
	< \delta_\ell
	\Bigr]
	>
	\prod_\ell (1-\kappa_\ell)
	>
	1-C_L.
	\label{eq: justification deviation bound}
\end{align}
Here, in the first inequality the dependence on $x$ disappears due to \eqref{eq: x dependence disappears}.

\subsubsection{Bound for deviations}\label{sec:__Design_A_bound_for_deviations}

We next consider the $\mathcal{S}_\ell$ for which we want to guarantee that
\begin{align}
	\mathbb{P}
	\bigl[
	\mathcal{S}_\ell
	<
	\delta_{\ell}
	\bigr]
	>
	1-\variable_{\ell}.
\end{align}
Let $\numberVar_\ell=\prod_{k=\ell}^L n_k$.
By assumption the $N^{(j_i)}_k$ are independent and identically $\mathrm{Normal}(0,\sigma^2_{k})$ distributed, where $\sigma^2_{k}\leq \sigma^2$, for some common $\sigma^2$.
We are lower bounding the number of copies required, therefore using \gls{AWGN} with higher variance only increases the lower bound, as the calculations below show.
We calculate the bound exemplary for $N^{(j_i)}$ distributed according to $\mathrm{Normal}(0,\sigma^2)$, re-substituting $\sigma_k^2$ below in \eqref{eq:final_step} (which is the bound given in \Cref{thm:recursive_case}) thus covers the case of $N^{(j_i)}_k \eqcom{d}= \mathrm{Normal}(0,\sigma^2_{k})$.

Each component of the vector
\begin{align}
	\sum_{j_i=1}^{n_{\ell-1}} N^{(j_i)} = \Bigl( \sum_{j_i=1}^{n_{\ell-1}} N^{(j_i)}_1,\dots, \sum_{j_i=1}^{n_{\ell-1}} N^{(j_i)}_{d_{\ell}} \Bigr)^\intercal
\end{align}
is assumed to be $\mathrm{Normal}(0,{ n_{\ell-1}}\sigma^2)=\sqrt{{ n_{\ell-1}}}\sigma\mathrm{Normal}(0,1)$ distributed.
It then holds that
\begin{align}
	\sum_{i=1}^{\numberVar_{\ell}}
	\Bigl\|
	\sum_{j_i=1}^{n_{\ell-1}}
	N^{(j_i)}
	\Bigr\|_2
	\eqcom{d}=
	\sum_{i=1}^{\numberVar_{\ell}}
	\sqrt{n_{\ell-1}}
	\sigma
	\|
	\mathrm{Normal}(0,I_{d})
	\|_2.
\end{align}

This is a sum of independent chi-distributed random variables, which means they are sub-gaussian (see below that we can calculate the sub-gaussian norm and it is indeed finite). Thus Hoeffding's inequality applies, according to which, for $X_1,\dots,X_n$ independent, mean zero, sub-gaussian random variables, for every $t\geq 0$
\begin{align}
	\mathbb{P}\Bigl[
		\Bigl| \sum_{i=1}^N X_i \Bigr|
		<
		t
	\Bigr]
	>
	1
	-
	2\exp
	\Bigl(
		-
		\frac{
			ct^2
		}{
			\sum_{i=1}^N \|X_i\|^2_{\psi_2}
		}
	\Bigr)
	\label{eq:Hoeffding}
\end{align}
holds;
see e.g.\ \cite[Theorem 2.6.2]{vershynin2018high}.
Here $c>0$ is an absolute constant (see \cite[Theorem 2.6.2]{vershynin2018high}) and
\begin{align}
	\|
	X
	\|_{\psi_2}
	:=
	\inf
	\bigl\{
	t>0
	:
	\mathbb{E}
	[\exp(
		X^2
		/
		t^2
	)]
	\leq
	2
	\bigr\}.
\end{align}
To apply Hoeffding's inequality in our setting, we need to center the occurring random variables.
For $N^{(i)}\sim \mathrm{Normal}(0,I_{d})$, the term $\|N^{(i)}\|_2$ is chi distributed with mean
\begin{align}
	\mu_d
	=
	\sqrt{2}
	\frac{
		\Gamma ((d+1)/2)
	}{
		\Gamma (d/2)},
	\label{eq:mean}
\end{align}
where $\Gamma$ is the gamma function, see e.g.\ \cite[p.238]{abell1999statistics}.

Consider
\begin{align}
&
\phantom{=}
	\mathbb{P}
	\Bigg[
	\frac{
		\prod_{i=\ell}^{L}a^{(i)}
		\prod_{i=\ell+1}^{L}\|W^{(i)}\|_\mathrm{op}
	}{
		\numberVar_\ell \sqrt{{ n_{\ell-1}}}
	}
	\sigma
	\sum_{i=1}^{\numberVar_{\ell}}
	\|
	N^{(i)}\|_2
	<
	\delta_{\ell}
	\Bigg]
\\
&
	=
	\mathbb{P}
	\Bigg[
	\frac{
		\prod_{i=\ell}^{L}a^{(i)}
		\prod_{i=\ell+1}^{L}\|W^{(i)}\|_\mathrm{op}
	}{
		\numberVar_\ell \sqrt{{ n_{\ell-1}}}
	}
	\sigma
	\sum_{i=1}^{\numberVar_{\ell}}
	\Bigl(\|
	N^{(i)}\|_2
	-
	\mu_{d_\ell}
	\Bigr)
\\
&
	\phantom{=\mathbb{P}\Bigg[}
	\qquad\quad
	<
	\delta_{\ell}
	-
	\frac{
		\prod_{i=\ell}^{L}a^{(i)}
		\prod_{i=\ell+1}^{L}\|W^{(i)}\|_\mathrm{op}
	}{
		\numberVar_\ell \sqrt{{ n_{\ell-1}}}
	}
	\sigma
	m_\ell
	\mu_{d_\ell}
	\Bigg]
\end{align}
which equals
\begin{align}
&
	\mathbb{P}
	\Bigg[
	\sum_{i=1}^{\numberVar_{\ell}}
	\Bigl(\|
	N^{(i)}\|_2
	-
	\mu
	\Bigr)
\\
&
\phantom{\mathbb{P}\Bigg[\sum_{i=1}^{\numberVar_{\ell}}}
	<
	\frac{
		\numberVar_\ell\sqrt{n_{\ell-1}}\delta_{\ell}
	}{
		\sigma
		\prod_{i=\ell}^{L}a^{(i)}
		\prod_{i=\ell+1}^{L}\|W^{(i)}\|_\mathrm{op}
	}
	-
	\numberVar_\ell
	\mu
	\Bigg]
\end{align}
and is lower bounded (compare to \eqref{eq:Hoeffding}) by
\begin{align}
	1
	-
	2\exp
	\left(\frac{
		-c
		\Bigl(
		\frac{
			\numberVar_\ell\sqrt{n_{\ell-1}}\delta_{\ell}
		}{
			\sigma
			\prod_{i=\ell}^{L}a^{(i)}
			\prod_{i=\ell+1}^{L}\|W^{(i)}\|_\mathrm{op}
		}
		-
		\numberVar_\ell
		\mu_{d_\ell}
		\Bigr)^2
	}{
		\sum_{i=1}^{\numberVar_\ell}
		\Big\|
		\|
		N^{(i)}\|_2
		-
		\mu
		\Big\|^2_{\psi_2}
	}\right),
	\label{key}
\end{align}
which in turn is lower bounded by
\begin{align}
	1
	-
	2\exp
	\left(\frac{
		-c
		\Bigl(
		\frac{
			\numberVar_\ell\sqrt{n_{\ell-1}}\delta_{\ell}
		}{
			\sigma
			\prod_{i=\ell}^{L}a^{(i)}
			\prod_{i=\ell+1}^{L}\|W^{(i)}\|_\mathrm{op}
		}
		-
		\numberVar_\ell
		\mu_{d_\ell}
		\Bigr)^2
	}{
		\sum_{i=1}^{\numberVar_\ell}
		C^2
		\Big\|
		\|
		N^{(i)}\|_2
		\Big\|^2_{\psi_2}
	}\right),
\end{align}
where $C>0$ is an absolute constant (see \cite[Lemma 2.6.8]{vershynin2018high}).
For a chi distributed random variable $\mathbf{X}$ it holds that
\begin{align}
	\mathbb{E}
	[\exp(
	\mathbf{X}^2
	/
	t^2
	)]
	=
	M_{\mathbf{X}^2}(1/t^2)
\end{align}
where $M_{\mathbf{X}^2}(s)$ is the moment generating function of $\mathbf{X}^2$---a chi-squared distributed random variable. It is known (see e.g.\ \cite[Appendix 13]{clapham2014concise}) that
\begin{align}
	M_{\mathbf{X}^2}(s)
	=
	(1-2s)^{-d_{\ell}/2}
\end{align}
for $s<\frac{1}{2}$. Accordingly for $2<t^2$, the property in the definition of the sub-gaussian norm
\begin{align}
	\mathbb{E}
	[\exp(
	\mathbf{X}^2
	/
	t^2
	)]
	=
	\Bigl(1-2\frac{1}{t^2}\Bigr)^{-d_{\ell}/2}
	\leq
	2
\end{align}
is satisfied for all $t$ for which 
\begin{align}
	t
	\geq
	\max
	\left\{
	\sqrt{
		\frac{4\sqrt[d_{\ell}]{4}}{2\sqrt[d_{\ell}]{4}-2}
	},
	\sqrt{2}
	\right\}
	=
	\sqrt{
		\frac{4\sqrt[d_{\ell}]{4}}{2\sqrt[d_{\ell}]{4}-2}
	}
\end{align}
holds. 
The square of the sub-gaussian norm of the chi distributed random variables is thus
\begin{align}
	\Big\|
	\|
	N^{(i)}\|_2
	\Big\|_{\psi_2}^2
	=
	\frac{4\sqrt[d_{\ell}]{4}}{2\sqrt[d_{\ell}]{4}-2}.
\end{align}
Substituting the norm into the lower bound yields
\begin{align}
	1
	-
	2\exp
	\left(\frac{
		-c
		\Bigl(
		\frac{
			\numberVar_\ell\sqrt{n_{\ell-1}}\delta_{\ell}
		}{
			\sigma
			\prod_{i=\ell}^{L}a^{(i)}
			\prod_{i=\ell+1}^{L}\|W^{(i)}\|_\mathrm{op}
		}
		-
		\numberVar_\ell
		\mu_{d_\ell}
		\Bigr)^2
	}{
		C^2
		\numberVar_\ell
		\frac{4\sqrt[d_{\ell}]{4}}{2\sqrt[d_{\ell}]{4}-2}
	}\right).
\end{align}

In order to achieve \eqref{eq: single layer condition}, a sufficient criterion is
\begin{align}
	\frac{\kappa_\ell}{2}
	\geq
	\exp
	\left(\frac{
		-c
		\numberVar_\ell
		\Bigl(
		\frac{
			\sqrt{n_{\ell-1}}\delta_{\ell}
		}{
			\sigma
			\prod_{i=\ell}^{L}a^{(i)}
			\prod_{i=\ell+1}^{L}\|W^{(i)}\|_\mathrm{op}
		}
		-
		\mu_{d_\ell}
		\Bigr)^2
	}{
		C^2
		\frac{4\sqrt[d_{\ell}]{4}}{2\sqrt[d_{\ell}]{4}-2}
	}\right).
\end{align}
Solving for $n_{\ell-1}$ leads to
\begin{align}
	n_{\ell-1}
&
	\geq
	\frac{
		\sigma^2
		\Bigl(\prod_{i=\ell}^{L}a^{(i)}
		\prod_{i=\ell+1}^{L}\|W^{(i)}\|_\mathrm{op}\Bigr)^2
	}{
		\delta_{\ell}^2
	}
\\
&
	\phantom{\geq}
	\times
	\Biggl(
	\sqrt{C^2
	\frac{4\sqrt[d_{\ell}]{4}}{2\sqrt[d_{\ell}]{4}-2}
	(-\ln\bigl(\kappa_\ell/2\bigr))
	\frac{1}{c
	\numberVar_\ell}
	}
	+
	\mu_{d_\ell}
	\Biggr)^2.
	\label{eq:final_step}
\end{align}
If we substitute the expression in \eqref{eq:mean} for $\mu_{d_\ell}$, \eqref{eq:final_step} becomes the bound as seen in \Cref{thm:recursive_case}.

\hfill$\square$

\subsection{Conclusion}
\label{sec:designA_discussion}

Within the context of the model described in \Cref{sec:Modeling_noise_in_ONNs}, we have established that any feed-forward \gls{NN} can be approximated arbitrarily well by \glspl{ONN} constructed using Design A. This is \Cref{thm:recursive_case} in essence.

This result has two consequences when it comes to the physical implementation of \glspl{ONN}.
On the one hand, it is guaranteed that the theoretical expressiveness of \glspl{NN} can be retained in practice.
On the other hand, Design A allows one to improve the accuracy of a noisy \gls{ONN} to a desired level, and in fact bring the accuracy arbitrarily close to that of any state-of-the-art feed-forward noiseless \glspl{NN}.
Let us finally remark that the high bandwidth of photonic circuits may be of use when implementing Design A.

\section{Results---Design B}
\label{sec:Results__Design_B}

\subsection{Reducing noise in feed-forward linear \texorpdfstring{\glspl{ONN}}{ONNs} (Design B)}
\label{sec:Results__Definition_of_Design_B}

Recall that an example of Design B is presented in \Cref{fig:Introduction__Schematic_depiction_of_the_two_noise_resistant_designs}(c).
\Cref{alg:Results__noise_reduction_tree_NN_Design_B} constructs this network, given a desired number of copies $m$ in each layer.

\begin{algorithm}
	\caption{Algorithm to construct a noise reducing network}
	\label{alg:Results__noise_reduction_tree_NN_Design_B}
	
	\begin{algorithmic}
		\REQUIRE Fix number $m \in \mathbb{N}$
		\REQUIRE $m$ copies of input $^1x^{(0)}, \dots, ^m x^{(0)}$
		\FOR{$\ell=1,\dots , L$}
		\FOR{$\alpha = 1,\dots , m$}
		\STATE $^{\alpha}\xi^{(\ell)} \gets W^{(\ell)} \,^\alpha x^{(\ell-1)} + b^{(\ell)} +\mathrm{Normal}(0,\Sigma_\text{w})$
		\ENDFOR
		\STATE $y^{(\ell)} \xleftarrow{\mathrm{combining}} \,^{1}\xi^{(\ell)} +\dots + ^{m}\xi^{(\ell)} $
		\STATE $ \left(^1y^{(\ell)}\dots,^my^{(\ell)}\right) \xleftarrow{\mathrm{splitting}} m^{-1}y^{(\ell)} $
		\FOR{$\alpha = 1,\dots , m$}
		\STATE $\,^\alpha x^{(\ell)} \gets \sigma^{(\ell)}(^\alpha y^{(\ell)})+\mathrm{Normal}(0,\Sigma_\text{act})$
		\ENDFOR
		\ENDFOR
		\RETURN $m^{-1} \sum_{\alpha = 1}^{m} \,^\alpha x^{(L)}$
	\end{algorithmic}
\end{algorithm}

Calculating the output of a \gls{NN} by using Design B first requires to fix a number $m$.
The input data $x^{(0)}$ is then modulated $m$ times, creating $m$ noisy realizations of the input $(^\alpha x^{(0)})_{\alpha = 1,\dots,m}$.
The weighted addition step and the activation function of each layer are singled out and copied $m$ times.
Both the copies of the weighted addition step and of the activation function of each layer are arrayed parallel to each other and performed on the $m$ inputs, resulting in $m$ outputs.
The $m$ parallel outputs of the weighted addition are merged to a single output,
and afterwards split into $m$ pieces.
The $m$ pieces are each send to one of the $m$ activation function mechanisms for processing.
The resulting $m$ activation values are the output of the layer.
If it is the last layer, the $m$ activation values are merged to produce the final output.
These steps are formally described in \Cref{alg:Results__noise_reduction_tree_NN_Design_B}.
A schematic representation of Design B can be seen in \Cref{fig:Introduction__Schematic_depiction_of_the_two_noise_resistant_designs}(c).

\subsection{Analysis of Design B}

We now consider the physical and mathematical consequences of Design B.

Observe that in Design B, the $m$ weighted additions of the $\ell$-th layer's input $x^{(l-1)}$ result in realizations $(^{\alpha}\xi^{(\ell)})_{\alpha =1,\dots,m}$ of $W^{(\ell)}x^{(\ell-1)}+b^{(\ell)}+\mathrm{Normal}(0,\Sigma_\text{w})$.
These realizations are then combined resulting in
\begin{equation}
  m W^{(\ell )} x^{(\ell -1)}
  +
  m b^{(\ell )}
  +
  \mathrm{Normal}(0,m\Sigma_\text{w})
  +
  \mathrm{Normal}(0,\Sigma_\text{sum})
  .
\end{equation}
Splitting the signal again into $m$ parts, each signal carries information following the distribution
\begin{align}
	W^{(\ell )} x^{(\ell -1)}
	+
	b^{(\ell )}
&
	+
	\mathrm{Normal}(0,m^{-1}\Sigma_\text{w}+m^{-2}\Sigma_\text{sum})
\\
&
	+
	\mathrm{Normal}(0,\Sigma_\text{spl})
	.
\end{align}
The mean of the normal distribution therefore is the original networks pre-activation obtained from this input (that is without perturbations).
The covariance matrix of the normal distribution is $m^{-1}\Sigma_\text{w}+m^{-2}\Sigma_\text{sum}+\Sigma_\text{spl}$.
Each of those signals is fed through the mechanism applying the activation function, yielding $m$ noisy versions of the output, distributed according to
\begin{align}
&
	x^{(\ell)} \mid x^{(\ell-1)}
\\
&
	\eqcom{d}=
	\sigma^{(\ell)}
	\bigl(
	W^{(\ell )} x^{(\ell -1)} + b^{(\ell )}
\\
&
	\phantom{\eqcom{d}=\sigma^{(\ell)}\bigl(}
	+ \mathrm{Normal}(0,m^{-1}\Sigma_\text{w}+m^{-2}\Sigma_\text{sum}+\Sigma_\text{spl})
	\bigr)
\\  &
	\phantom{\eqcom{d}=}
	+
	 \mathrm{Normal}(0,\Sigma_\text{act})
	 .
\end{align}

The effect of Design B is thus that $T^{(\ell)}( \Sigma )$ in \eqref{eq:Model__definition_iteration_map} is replaced by
\begin{align}
  T^{(\ell)}_m( \Sigma )
   &
  :=
  \frac{1}{m}D^{(\ell)}W^{(\ell)}\Sigma \bigl(D^{(\ell)}W^{(\ell)}\bigr)^\intercal
  +
  \frac{1}{m}D^{(\ell)} \Sigma_\text{w} \bigl(D^{(\ell)}\bigr)^\intercal
  \nonumber \\  &
     \phantom{:=}
     +
     \frac{1}{m^2}D^{(\ell)} \Sigma_\mathrm{sum} \bigl(D^{(\ell)}\bigr)^\intercal
     +
     D^{(\ell)} \Sigma_\mathrm{spl} \bigl(D^{(\ell)}\bigr)^\intercal
     +
     \Sigma_\mathrm{a}
     ;
\end{align}
see \Cref{sec:Results__Explanation of the effect of Design B}.
Observe also that $\Sigma_a^{(\ell)}$ can be written as $(1/m) m \Sigma_a^{(\ell)}$.
Therefore, if we substitute the matrix $\bar{\Sigma}_a^{(\ell)} = m\Sigma_a^{(\ell)}$ for $\Sigma_a^{(\ell)}$ in $T^{(\ell)}( \Sigma )$, we can write
\begin{align}
  T_m^{(\ell)}( \Sigma )
   &
  =
  {m}^{-1} T^{(\ell)}( \Sigma )+
  {m}^{-2}D^{(\ell)} \Sigma_\mathrm{sum} \bigl(D^{(\ell)}\bigr)^\intercal
  \nonumber \\  &
     \phantom{=}
     +
     D^{(\ell)} \Sigma_\mathrm{spl} \bigl(D^{(\ell)}\bigr)^\intercal
     .
\end{align}

We have the following analogs to \Cref{prop:Model__recusive_structure_of_covariances} and \Cref{cor:Limit_for_symmetric_linear_ONNs}:

\begin{theorem}[Distribution of Design B]
  \label{thm:Results__DesignB_recusive_structure_of_covariances}

  Assume that there exist vectors $a^{(\ell)}\in \R^{d_\ell}$ such that $\sigma^{(\ell)}(y) = \textrm{diag}(a^{(\ell)}) y$.
  The feed-forward linear \gls{ONN} constructed using Design B with $m$ copies then satisfies
  \begin{equation}
    \Psi_m^{\mathrm{ONN}}(\cdot, w)
    \eqcom{d}=
    \mathrm{Normal}
    \bigl(
    \Psi^{\mathrm{NN}}(\cdot, w)
    ,
    \Sigma_{\mathrm{ONN},m}^{(L)}
    \bigr)
    ,
  \end{equation}
  where for $\ell = L, L-1, \ldots, 1$,
  \begin{equation}
    \Sigma_{\mathrm{ONN},m}^{(\ell)}
    =
    T_m^{(\ell)}( \Sigma_{\mathrm{ONN},m}^{(\ell-1)} )
    ;
    \quad
    \textnormal{and}
    \quad
    \Sigma_{\mathrm{ONN},m}^{(0)}
    = \Sigma_{\mathrm{m}}
    .
  \end{equation}
\end{theorem}

Under the assumption of symmetric noise, a similar simplification of the recursion in \Cref{thm:Results__DesignB_recusive_structure_of_covariances}, similar to that in \Cref{prop:Model__recusive_structure_of_covariances}, is possible.
Assume $\Sigma_\mathrm{sum}=\Sigma_\mathrm{spl}=0$.
Introduce again $P^{(\ell)} := \prod_{i=\ell+1}^LD^{(i)} W^{(i)}$ for notational convenience.
The following is proved in \Cref{sec:Appendix__proofs_of_DesignB}:

\begin{corollary}[Symmetric noise case]
  \label{cor:Results__DesignB_Limit_for_symmetric_linear_ONNs}

  Assume that for all $\ell \in \naturalNumbersPlus$,
  $\Sigma_{\mathrm{a}}^{(\ell)} = \Sigma_{\mathrm{a}}$
  and
  $\Sigma_{\mathrm{w}}^{(\ell)} = \Sigma_{\mathrm{w}}$.
  Then,
  \begin{align}
    \Sigma_{\mathrm{ONN},m}^{(L)}
     & =
    \sum_{\ell=1}^L (m^{-1})^{L-\ell} P^{(\ell)} m\Sigma_\mathrm{a} (P^{(\ell)})^\intercal
    \nonumber \\  & \phantom{=}
       +
       \sum_{\ell=1}^L
       (m^{-1})^{L-\ell}
       P^{(\ell)}
       D^{(\ell)}\Sigma_\mathrm{w} (D^{(\ell)})^\intercal
       \bigl(P^{(\ell)}\bigr)^\intercal
    \nonumber \\  & \phantom{=}
       +
       (m^{-L})
       P^{(0)}
       \Sigma_\mathrm{m}
       ( P^{(0)} )^\intercal.
  \end{align}
\end{corollary}

We will next consider the limit of the covariance matrix in a large, symmetric linear \gls{ONN} with Design B, that we can grow infinitely deep.
\Cref{alg:Results__noise_reduction_tree_NN_Design_B} is namely able to guarantee boundedness of the covariance matrix in such deep \gls{ONN} if the parameter $m$ is chosen appropriately:

\begin{corollary}
  \label{cor:optimal_scaling_linear_case}

  Consider a linear \gls{ONN} with Design B and parameter $m$, that has $L$ layers, and that satisfies the following symmetry properties:
  for all $\ell \in \{ 1, \ldots, L \}$,
  $W^{(\ell)} = W$,
  $D^{(\ell)} = D$,
  $\Sigma_{\mathrm{a}}^{(\ell)} = \Sigma_{\mathrm{a}}$
  and
  $\Sigma_{\mathrm{w}}^{(\ell)} = \Sigma_{\mathrm{w}}$.
  Then, if $\pnorm{ D }{\mathrm{F}}\pnorm{W}{\mathrm{F}}<\sqrt{m}$, the limit $\lim_{L \to \infty} \Sigma^{(L)}_{\mathrm{ONN}}$ exists.
  
  Moreover,
  \begin{align}
  	& \lim_{L \to \infty}
  	\Sigma^{(L)}_{\mathrm{ONN},m} \nonumber                                                                                                   \\
  	& = \sum_{n=0}^\infty m^{-(n+1)} (DW)^n \left(D \Sigma_\mathrm{w} D^\intercal + m\Sigma_\mathrm{a}\right) \left((DW)^n\right)^\intercal.
  	\label{eq:Results__Corollary2}
  \end{align}
\end{corollary}

Notice that the bound on the number of copies needed for the covariance matrix of an \gls{ONN} to converge to a limit is independent of e.g.\ the Frobenius norms of the covariance matrices that describe the noise distributions.
This is because, here, we are not interested in bounding the covariance matrix to a specific level; instead, we are merely interested in the existence of a limit.

\subsection{Discussion \& Conclusion}

Compared to \Cref{thm:Results__DesignB_recusive_structure_of_covariances}'s recursive description of the covariance matrix in any linear \gls{ONN} with Design B, \Cref{cor:Results__DesignB_Limit_for_symmetric_linear_ONNs} provides a series that describes the covariance matrix in any linear, symmetric \gls{ONN} with Design B.
While the result holds more restrictively, it is more insightful.
For example, it allows us to consider the limit of the covariance matrix in an extremely deep \glspl{ONN} (see \Cref{cor:optimal_scaling_linear_case}).
\Cref{cor:optimal_scaling_linear_case} suggests that in deep \glspl{ONN} with Design B, one should choose
$
  m
  \approx
  \lceil
  ( \pnorm{ D }{\mathrm{F}}\pnorm{W}{\mathrm{F}} )^2
  \rceil
$
in order to control the noise and not be too inefficient with the number of copies.

These results essentially mean that in a physical implementation of an increasingly deep and linear \gls{ONN}, the covariance matrix can be reduced (and thus remain bounded) by applying Design B with multiple copies.
The quality of the \gls{ONN}'s output increases as the number of copies in Design B (or Design A for that matter) is increased.
Finally, it is worth mentioning that Design B could potentially be implemented such that it leverages the enormous bandwidth of optics.

\section{Simulations}
\label{sec:Numerical_analysis}
We investigate the improvements of output quality achieved by Designs A and B on a benchmark example: the convolutional neural network LeNet \cite{lecun1998gradient}.
As measure for quality we consider the \gls{MSE}
\begin{align}
  \mathrm{MSE} = \frac{1}{n}\sum_{i=1}^{n} \left(\Psi^{\mathrm{ONN}}(x^{(i)}, w)
  -
  \Psi^{\mathrm{NN}}(x^{(i)}, w) \right)^2
\end{align}
and the prediction accuracy
\begin{align}
  \frac{\# \{ \mathrm{correctly~classified~images}\}}{\# \{ \mathrm{images}\}}.
\end{align}

\subsection{Empirical variance}

We extracted plausible values for $\Sigma_\mathrm{w}$ and $\Sigma_\mathrm{a}$ from the \gls{ONN} implementation \cite{shi2022inp} of a $2$-layer \gls{NN} for classification of the \gls{MNIST} database \cite{deng2012mnist}.
In \cite{shi2022inp}, the authors trained the \gls{NN} on a classical computer and implemented the trained weights afterwards in an \gls{ONN}.
They then tuned noise (with the same noise model as in \Cref{sec:Modeling_noise_in_ONNs} of this paper) into the noiseless computer model, assuming that $\Sigma_\mathrm{w}=\mathrm{diag}(\sigma_\mathrm{w}^2)$ and $\Sigma_\mathrm{a}=\mathrm{diag}(\sigma_\mathrm{a}^2)$.
They found $\sigma_\mathrm{w}\in [0.08,0.1]\cdot d$ and $ \sigma_\mathrm{a}\in [0.1,0.15]\cdot d$ to reach the same accuracy levels as the \gls{ONN}, where $d$ denotes the diameter of the range.

\subsection{LeNet \texorpdfstring{\gls{ONN}}{ONN}: Performance when implemented via Design A and B}
\label{Section A versus B}

Convolutional \glspl{NN} can be regarded as feedforward \glspl{NN} by stacking the (2D or 3D) images into column vectors and arranging the filters to a weight matrix.
Thus Design A and B are well-defined for Convolutional \glspl{NN}.
We apply the designs to LeNet5 \cite{lecun1998gradient}, which is trained for classifying the handwritten digits in the \gls{MNIST} dataset \cite{deng2012mnist}.
The layers are:
\begin{itemize}
  \item[1.]
    2D convolutional layer with kernel size $5$, stride $1$ and $2$-padding.
    Output has $6$ channels of $28 \mathrm{x} 28$ pixel representations, with the activation function being $\tanh$;
  \item[2.]
    average pooling layer, pooling $2 \mathrm{x} 2$ block, the output therefore is $14 \mathrm{x} 14$;
  \item[3.]
    2D convolutional layer with kernel size $5$, stride $1$ and no padding.
    The output has $16$ channels of $10 \mathrm{x} 10$ pixel representations and the activation function is $\tanh$;
  \item[4.]
    average pooling layer, pooling $2 \mathrm{x} 2$ block, the output therefore is $5 \mathrm{x} 5$;
  \item[5.]
    2D convolutional layer with kernel size $5$, stride $1$ and no padding.
    The output has $120$ channels of $1$ pixel representations and the activation function used is $\tanh$;
  \item[(5.)]
    flattening layer, which turns the $120$ one-dimensional channels into one $120$-dimensional vector;
  \item[6.]
    dense layer with $84$ neurons and $\tanh$ activation function;
  \item[7.]
    dense layer with $10$ neurons and $\mathrm{softmax} $ activation function.
\end{itemize}

\begin{figure}[t]
  \centering
  \includegraphics[width = 0.95\linewidth]{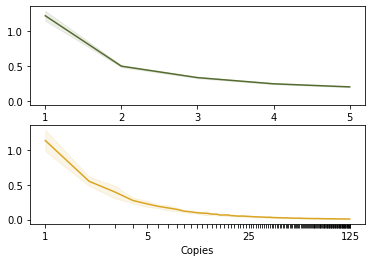}
  \caption{
    \gls{MSE}($\cdot 10^2$) for Design A (top) and Design B (bottom) as function of copies on LeNet5 trained for \gls{MNIST} classification.
    The pale area contains the 95\%-confidence intervals.
  }
  \label{fig: LeNet_MSE_AvsB}
\end{figure}

\begin{figure}[hbtp]
  \centering
  \includegraphics[width = 0.95\linewidth]{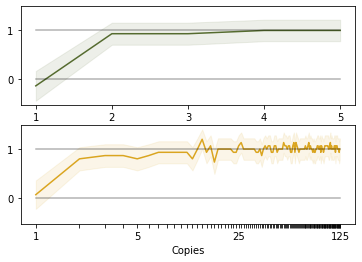}
  \caption{
    Relative accuracy for Design A (top) and Design B (bottom) as function of copies on LeNet5 trained for \gls{MNIST} classification.
    The pale area contains the $56.5$\%-confidence intervals.
  }
  \label{fig: LeNet_acc_AvsB}
\end{figure}

\Cref{fig: LeNet_MSE_AvsB,fig: LeNet_acc_AvsB} show the \gls{MSE} and the prediction accuracy of Design A and B for an increasing number of copies, respectively.

For simplicity we set all individual \emph{copies} $n_i$ \emph{per layer} $i$ in Design A to equal $m$, that is $n_i=m$ for all $i$. The total number of copies that Design A starts with then is $m^L$. Here $L$ is equal to $7$.
In Design B the number of copies is $m$ \emph{per layer} and the total number of copies is $mL$.
In the case of one copy the designs A and B are identical to the original network, while we focus on the effect once the designs deviate from the original network ($m\geq 2$).

The axis in \Cref{fig: LeNet_MSE_AvsB,fig: LeNet_acc_AvsB} denote the number of copies \emph{per layer}. Here, we scale the copies per layer for Design A linearly, because the total amount of copies for Design A grows exponentially and we scale the copies per layer for Design B exponentially, because the total number of copies for Design B grows linearly. This way the comparison is on equal terms.

\Cref{fig: LeNet_MSE_AvsB} displays the \gls{MSE} seen for LeNet, depending on the amount of copies for each design.
In the trade-off between additional resources needed for the additional copies against the diminishing benefits of adding further copies, we see that, for both measures \gls{MSE} (\Cref{fig: LeNet_MSE_AvsB}) and relative accuracy (\Cref{fig: LeNet_acc_AvsB}), already 2 to 5 copies per layer yield good results.
The relative accuracy in \Cref{fig: LeNet_acc_AvsB} is scaled such that $0$ corresponds to the accuracy of the original \gls{NN} with noise profile (i.e., the \gls{ONN} without modifications, we call this the original \gls{ONN}) and $1$ to the accuracy of the original \gls{NN} without noise.
The designs do not alter the fundamental operation of the original \gls{NN}, therefore there should be no performance gain and the original \gls{NN}'s accuracy should be considered the highest achievable, thus constituting the upper bound in relative accuracy of $1$.
Likewise the lowest accuracy should be given by the original \gls{ONN}, as there is no noise reduction involved.

\subsection{Effect of additional layers in LeNet}
\label{sec:Effect_of_additional_layers_in_LeNet}

In order to investigate how the depth affects the noise at the output, while keeping the operation of the network the same to ensure the results are commensurable, we insert additional layers with identity matrix and identity activation function (we will call them identity layers) into a network.
Specifically, we take networks with the LeNet architecture as in \Cref{Section A versus B}, using different activation functions, while fixing the output layer to be $\mathrm{softmax}$.
We then insert identity layers between layers 1 and 2, 3 and 4, 5 and 6, as well as between layers 6 and 7.
For a fixed total of additional layers, the layers are inserted in the four spots between layers ${1\&2}$, ${3\&4}$, ${5\&6}$, and ${7\&8}$ according to the tuple
\begin{align}
  n\mapsto \left( \Bigl\lfloor \frac{n+3}{4} \Bigl\rfloor, \Bigl\lfloor \frac{n+2}{4} \Bigl\rfloor, \Bigr\lfloor \frac{n+1}{4} \Bigr\rfloor, \Bigr\lfloor \frac{n}{4} \Bigr\rfloor \right).
\end{align}
The insertion pattern is illustrated in \Cref{table:insertions}:

\begin{table}[hbtp]
  \centering
  \scriptsize
  \begin{tabular}{ccccc}
    \toprule
    \# of additional layers & {1\&2} & {3\&4} & {5\&6} & {7\&8} \\
    \midrule
    1                       & 1      & 0      & 0      & 0      \\
    2                       & 1      & 1      & 0      & 0      \\
    3                       & 1      & 1      & 1      & 0      \\
    4                       & 1      & 1      & 1      & 1      \\
    5                       & 2      & 1      & 1      & 1      \\
    6                       & 2      & 2      & 1      & 1      \\
    \ldots                  & \ldots & \ldots & \ldots & \ldots \\
    \bottomrule
  \end{tabular}
  \caption{Insertion pattern.
  }
  \label{table:insertions}
\end{table}

\begin{figure*}
  \centering
  \includegraphics[width=\linewidth]{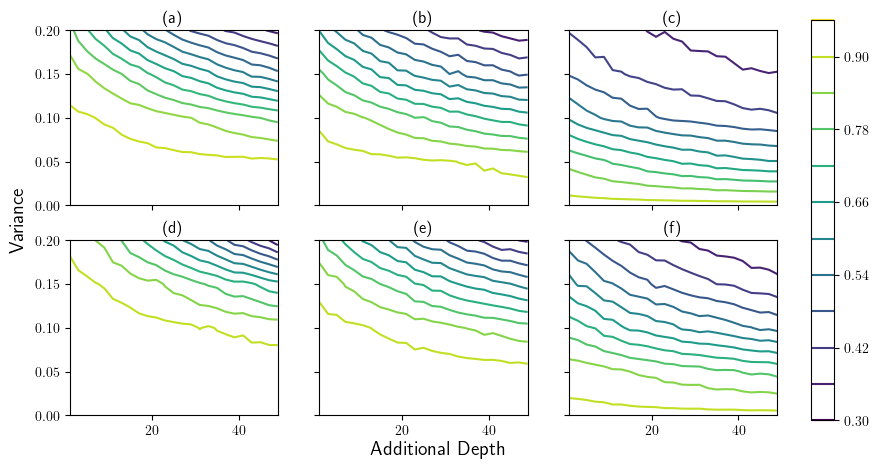}
  \caption{
    Accuracy of LeNet \glspl{ONN}, depending on the amount of inserted identity layers and the variance level of the \gls{ONN}, for
    (a) a network with $\tanh$ activation function and one copy,
    (b) a network with  $\mathrm{ReLU}$ activation function and one copy,
    (c) a network with  $\mathrm{linear}$ activation function and one copy,
    (d)	a network with $\tanh$ activation function and two copies,
    (e) a network with  $\mathrm{ReLU}$ activation function and two copies,
    (f) a network with  $\mathrm{linear}$ activation function and two copies.
  }
  \label{fig:LeNet_different_act_fcts_copies_1}
\end{figure*}

Finally, we tune the variance terms of the covariance matrix in our noise model.
The results are displayed in \Cref{fig:LeNet_different_act_fcts_copies_1}.

In \Cref{fig:LeNet_different_act_fcts_copies_1}, we observe that the $\tanh$ and the $\mathrm{ReLU}$ networks perform as expected.
Additional noisy layers decrease the accuracy and thus the same level of performance can only be achieved if the variance is lower.
This trend can also be seen in the linear network, but to a lesser extend.

\subsection{Simulations on effective values for Design B}
\label{Section Sim Design B}

\begin{figure}
  \includegraphics[width = 0.95\linewidth]{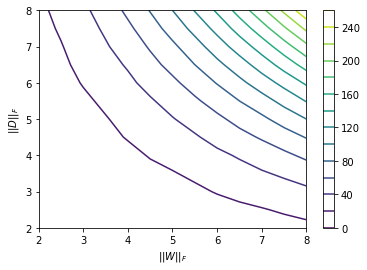}

  \caption{The contour lines denote the lowest $m$ for which $\pnorm{\Sigma_{\mathrm{ONN},m}^{(L)}}{\mathrm{F}}$ stops growing exponentially, as a function of $\pnorm{W}{\mathrm{F}}$ and $\pnorm{D}{\mathrm{F}}$.}
  \label{fig:m__normD_and_normW}
\end{figure}

According to \Cref{cor:optimal_scaling_linear_case}, the covariance matrix of a linear \gls{ONN} constructed by Design B is bounded if $m > \allowbreak \left(\pnorm{ D }{\mathrm{F}}\pnorm{W}{\mathrm{F}}\right)^2$, and therefore
$
  m
  =
  \lceil \left(\pnorm{ D }{\mathrm{F}}\pnorm{W}{\mathrm{F}}\right)^2
  \rceil
$,
is sufficient to ensure that the covariance matrix of the output distribution
$
  \Psi_m^{\mathrm{ONN}}(\cdot, w)
  \eqcom{d}=
  \mathrm{Normal}
  (
  \Psi^{\mathrm{NN}}(\cdot, w)
  ,
  \Sigma_{\mathrm{ONN},m}^{(L)}
  )
$
in \Cref{thm:Results__DesignB_recusive_structure_of_covariances} is bounded in linear \glspl{NN}.
This is derived by using submultiplicativity of the norm (see \eqref{eq:appendix__tensor_product_goes_to_zero_m}) and is therefore possibly a lose bound.
We use the exact relation given by \Cref{cor:Results__DesignB_Limit_for_symmetric_linear_ONNs} for the covariance matrix in \Cref{thm:Results__DesignB_recusive_structure_of_covariances} to investigate the lowest values for $m$ for which the covariance matrix starts being bounded.
In \Cref{fig:m__normD_and_normW} we depict a linear \gls{NN} with constant width $4$. We vary the values for $\pnorm{D}{\mathrm{F}}$ and $\pnorm{W}{\mathrm{F}}$.
Upon close inspection we see that the lowest value for $m$ seems to be
$
  g(x, y)
  \simeq
  \lceil
  ( x y )^2 / \pnorm{I_d}{\mathrm{F}}^4
  \rceil
  ,
$
where $I_d$ is the identity matrix of dimension $d$, see \Cref{fig:m__normD_and_normW}.
Because $\pnorm{I_d}{\mathrm{F}}=\sqrt{d}$, the value for $m$ found numerically is
$
  m
  \simeq
  \lceil
  ( \pnorm{ D }{\mathrm{F}}\pnorm{W}{\mathrm{F}} / d )^2
  \rceil
  .
$

\section{Discussion \& Conclusion}
\label{sec:Discussion}

Design A, introduced in \Cref{sec:Results__Design_A}, guarantees an approximation property (\Cref{thm:recursive_case}).
This is achieved through technical machinery to control the noise, even though there are nonlinear activation functions involved.
This method is powerful enough to yield the universal approximation property, as \glspl{NN} can be approximated arbitrarily well with \glspl{ONN} that are constructed through the first design, and \glspl{NN} themselves can approximate any continuous function arbitrarily well \cite[Theorem 1]{leshno1993multilayer}.
Our mathematical guarantee however, only states a sufficient number of copies required, and this number grows exponentially as the number of layers increases.

We then introduced Design B in \Cref{sec:Results__Design_B}, in which the growth of number of copies is much more benign.
However, the analysis of Design B was restricted to linear \glspl{NN}, and Design B might therefore not be expressive enough to have the universal approximation property.
Linear \glspl{NN}, or \glspl{NN} with algebraic polynomials as activation functions for that matter, namely do not posses the universal approximation property.
The assumption of linear activation functions did allow us to characterize the distribution of the output exactly on the flipside (\Cref{thm:Results__DesignB_recusive_structure_of_covariances}).

In short, in this paper, we have discussed the noise present in \glspl{ONN} and described a mathematical model for the noise.
We also investigate the numerical implications of the the mathematical model, with a specific focus on the effects of depth (\Cref{fig:LeNet_different_act_fcts_copies_1}).
The proposed noise reduction schemes yield greater accuracy and the theoretical results (\Cref{thm:recursive_case} and \Cref{cor:optimal_scaling_linear_case}) guarantee that \glspl{ONN} work just as noiseless \glspl{NN} in the many copies limit.
With the designs and findings of \Crefrange{sec:Results__Design_A}{sec:Results__Design_B} we have a framework to exploit known \gls{NN} wisdom, as no new training is required.
Further research should address optimization algorithms that take the noise of \glspl{ONN} into account to investigate the regularization, generalization and minimization properties of trained \glspl{ONN}.

\section*{Acknowledgments}
This research was supported by the European Union's Horizon 2020 research and innovation programme under the Marie Skłodowska-Curie grant agreement no. 945045, and by the NWO Gravitation project NETWORKS under grant no. 024.002.003.
\inlinegraphics{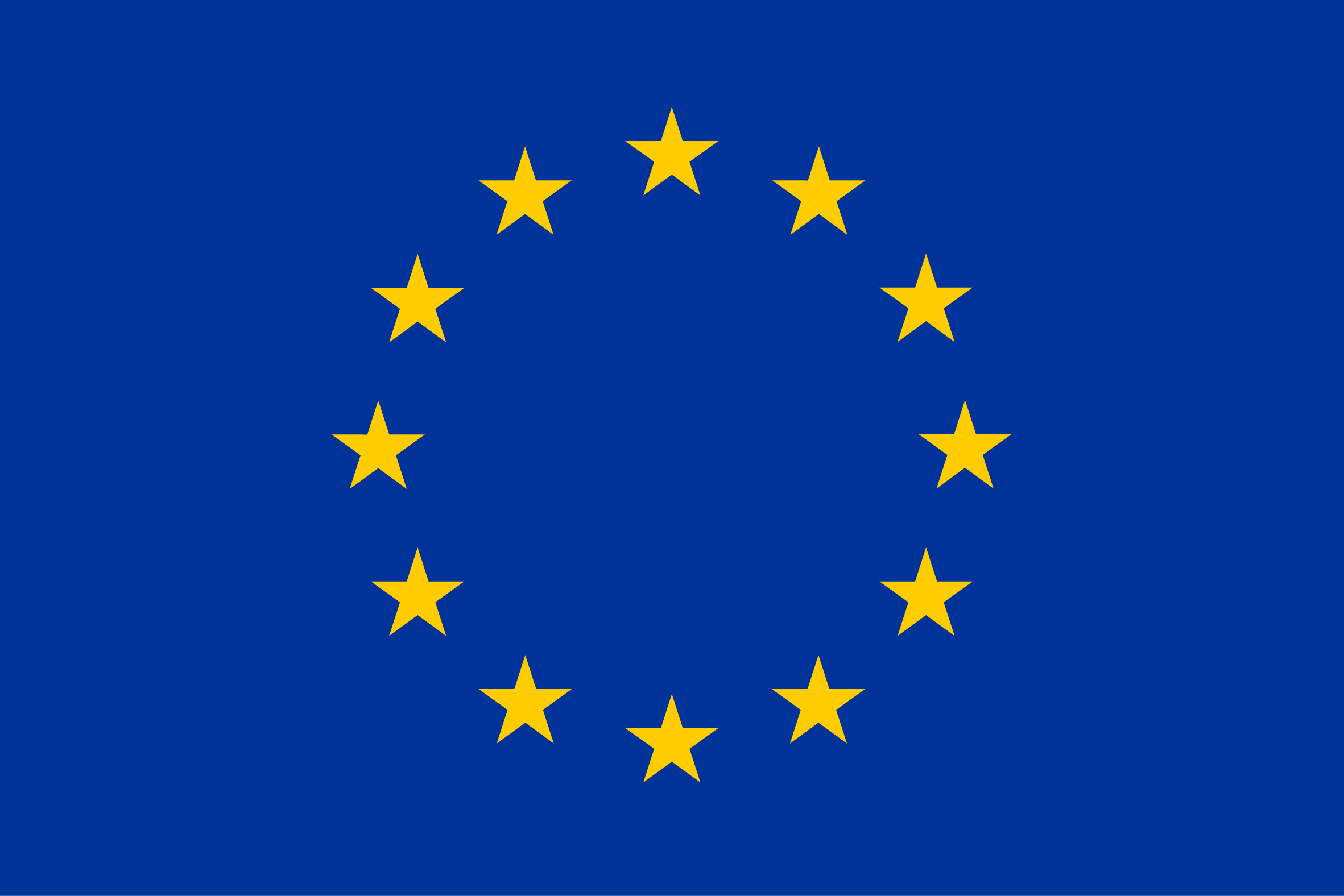}

We would finally like to thank Bin Shi for advise on the noise level parameters of \glspl{ONN} for our simulations.
Furthermore we want to thank Albert Senen--Cerda, Sanne van Kempen, and Alexander Van Werde for feedback to a draft version of this document.

\bibliography{sections/Bib}{}
\bibliographystyle{elsarticle-num}

\vfill

\appendix
\section{}

\subsection{Proofs of \Cref{sec:Modeling_noise_in_ONNs}}

\subsubsection{Proof of \Cref{cor:Limit_for_symmetric_linear_ONNs}}
\label{sec:Limit_for_symmetric_linear_ONNs}

The first conclusion of \Cref{cor:Limit_for_symmetric_linear_ONNs} follows immediately from expanding the recursion.

Assume that for all $\ell \in \mathbb{N}$ $\mathrm{diag}(a^{(\ell)} )=a$ and $W^{(\ell)}=W$ as well as $\Sigma_\mathrm{a}^{(\ell)} = \Sigma_\mathrm{a}$ and $\Sigma_\mathrm{w}^{(\ell)} = \Sigma_\mathrm{w}$.
Let $T:=T^{(\ell)}$ be the common map under those conditions.

To prove the second conclusion of \Cref{cor:Limit_for_symmetric_linear_ONNs}, observe that for any two matrices $X$ and $Y$ of the same dimension as $a$ and $W$,
\begin{align}
  \| T(X)-T(Y) \|_F
   &
  =
  \| a W (X - Y) (a W)^\intercal \|_F
  \\
   &
  \leq
  \|a\|_F^2 \|W\|_F^2 \|X-Y\|_F
  \label{eq:Results__contracting_prop}
\end{align}
by submultiplicativity of the Frobenius norm.
Let us now consider the setting of \Cref{prop:Model__recusive_structure_of_covariances} for a moment, that is, we initialize $X_1=\Sigma_{\mathrm{m}}$ and calculate $X_{\ell +1} = T(X_\ell)$ recursively.
The sequence $X_1, X_2, \ldots$ converges if $\|a\|_F \|W\|_F <1$, as a consequence of the Banach fixed point theorem \cite{banach1922operations}
combined with \eqref{eq:Results__contracting_prop}.

We may therefore consider the unique fixed point
$
  X_\star := \lim_{\ell\rightarrow \infty} X_\ell.
$
It must satisfy the fixed point equation $T(X_\star) = X_\star$, which reads
$
  X_\star - (aW) X_\star (aW)^\intercal = a \Sigma_\mathrm{w} a^\intercal + \Sigma_\mathrm{a}
  .
$
Equivalently,
\begin{align}
  \mathrm{vec} (X_\star) & - ((aW) \otimes (aW)) \mathrm{vec}(X_\star)
  \\
                         &
  = \mathrm{vec}(a \Sigma_\mathrm{w} a^\intercal) + \mathrm{vec}(\Sigma_\mathrm{a})
  .
\end{align}
Here, $\otimes$ denotes the Kronecker product and $\mathrm{vec}$ the vectorization of a matrix (effectively, we stack the columns of the matrix $A$ on top of one another).
This vectorization trick allows us to write the solution to the fixed point equation as
\begin{align}
  \mathrm{vec} (X_\star)
  =
  ( I - ((aW)^{\otimes 2}) )^{-1} (\mathrm{vec}(a \Sigma_\mathrm{w} a^\intercal + \Sigma_\mathrm{a}))
  .
  \label{eq:Results__solution_fixed point equation_before_neumann}
\end{align}
Here, $(aW)^{\otimes 2}$ denotes $(aW)\otimes (aW)$.

Formally we rewrite the inverse in \eqref{eq:Results__solution_fixed point equation_before_neumann} in terms of a von Neumann series
\begin{align}
  ( I - ((aW)^{\otimes 2}) )^{-1}
  =
  \sum_{n=0}^\infty ((aW)^{\otimes 2})^n
  .
  \label{eq:Results__neumann_in_fixedpointeq}
\end{align}
This is however justified only if
\begin{align}
  \| ((aW)^{\otimes 2})^n \|_F
  \rightarrow
  0
  ,
  \label{eq:Results__tensor_product_goes_to_zero}
\end{align}
which we verify next.

For the Kronecker product it holds that $\mathrm{tr}(A\otimes B)=\mathrm{tr}(A)\mathrm{tr}(B)$.
Therefore, $\| (aW)^{\otimes 2} \|_F = \mathrm{tr}( (aW)^\intercal (aW) ) = \| aW \|_F^2$ by definition of the Frobenius norm.
Furthermore, by submultiplicativity,
$
  \| ((aW)^{\otimes 2})^n \|_F
  \leq
  \| (aW)^{\otimes 2} \|_F^n
  .
$
Thus, by the assumption that $\|a\|_F \|W\|_F < 1$, condition \eqref{eq:Results__tensor_product_goes_to_zero} holds and \eqref{eq:Results__neumann_in_fixedpointeq}'s expression is proper.
This leads to the representation of $X_\ast$ as
\begin{align}
   &
  \mathrm{vec} (X_\star)
  =
  \sum_{n=0}^\infty
  ((aW)^{\otimes 2})^n \mathrm{vec}(a \Sigma_\mathrm{w} a^\intercal + \Sigma_\mathrm{a} )
  \nonumber \\  &
     =
     \mathrm{vec}( a \Sigma_\mathrm{w} a^\intercal + \Sigma_\mathrm{a} )
     + \sum_{n=1}^\infty
     ((aW)^{\otimes 2})^n \mathrm{vec}(a \Sigma_\mathrm{w} a^\intercal + \Sigma_\mathrm{a} )
     .
     \label{eq:Results__solution_fixed point equation_with_neumann}
\end{align}
Returning to matrix notation we have
\begin{equation}
  X_\star
  =
  a \Sigma_\mathrm{w} a^\intercal + \Sigma_\mathrm{a}
  + \sum_{n=1}^\infty (aW)^n (a \Sigma_\mathrm{w} a^\intercal + \Sigma_\mathrm{a}) ((aW)^n)^\intercal
  .
  \label{eq:Results__solution_fixpoint}
\end{equation}
This proves the second conclusion of \Cref{cor:Limit_for_symmetric_linear_ONNs}.

\subsection{Additional material \Cref{sec:Results__Design_B}}\label{sec:Appendix__proofs_of_DesignB}

\subsubsection{Additional considerations on Design A}\label{appendix: Design A}
To consider the total number of copies in Design A to guarantee  \eqref{eq:DesignA_desired_prop}, we need to multiply all the $n_\ell$ in \Cref{thm:recursive_case}. To simplify the terms we upper bound
\begin{align}
	\sqrt{2}
	\frac{
			\Gamma ((d_{\ell+1}+1)/2)
		}{
			\Gamma (d_{\ell+1}/2)
		}
	\qquad
	\ell\in
	\{
	0,\dots,L-1
	\}
\end{align}
by a constant $D$ (assuming the sequence of $d_\ell$ is bounded). We also replace
\begin{align}
	\sqrt{
			\frac{C^2}{c}
			\frac{4\sqrt[d_{\ell+1}]{4}}{2\sqrt[d_{\ell+1}]{4}-2}
			(-\ln\bigl(\kappa_{\ell+1}/2\bigr))
			\frac{1}{
					\prod_{i=\ell +1}^L n_i}
		}
\end{align}
by a constant $E$.
If the total number of copies satisfies
\begin{align}
	\prod_{\ell=0}^{L}
	n_{\ell}
	&\geq
	\frac{
			\Big(\prod_{\ell=1}^{L}\big(a^{(\ell)}\big)^{2\ell}\Big)
			\Big(
			\prod_{\ell=1}^{L}\big(\|W^{(\ell)}\|_\mathrm{op}\big)^{2(\ell-1)}
			\Big)
		}{
			\prod_{\ell=1}^{L}\delta_{\ell}^2
	}
	\\
	&
	\phantom{\geq}
	\times
	\sigma^{2L}
	\bigl(
	D
	+
	E
	\bigr)^{2L},
\end{align}
then we are able to construct an \gls{ONN} $\tilde{\Psi}^{\mathrm{ONN}}$ that satisfies \eqref{eq:DesignA_desired_prop}. The product $\prod_{\ell=1}^{L}\delta_{\ell}^2$ is maximized if all $\delta_\ell=\mathcal{D}_L/L$. We furthermore upper-bound
$\prod_{\ell=1}^{L}\big(a^{(\ell)}\big)^{2\ell}$ by $a^{2L^2}$ and
$\prod_{\ell=1}^{L}\big(\|W^{(\ell)}\|_\mathrm{op}\big)^{2(\ell-1)}$ by $W^{2L^2}$. We then have
\begin{align}
	N = \prod_{\ell=0}^{L}
	n_{\ell} =\omega(  K^{2L+2L^2} L^L )=\omega(  K^{2L(L+1)} L^L ).
\end{align}

\subsubsection{Deriving the covariance matrix for Design B}\label{sec:Results__Explanation of the effect of Design B}

We now derive $T_m^{(\ell)}(\Sigma)$---the transformation of the covariance matrix which an input undergoes as it becomes the output of layer $\ell$.
Recall that this input is distributed as $\mathrm{Normal}( x^{(\ell-1)} , \allowbreak \Sigma )$.
Denote the random variable for the pre-activation (from which the realizations are drawn) after joining and splitting beams by $\mathbf{P}^{(\ell)}$.
Then
\begin{align}
   &
  \mathbf{P}^{(\ell)}
  \mid
  x^{(\ell-1)}
\\ &
  \eqcom{d}= m^{-1}
  \Bigl[ \bigl(\sum_{i=1}^m W^{(\ell)} \mathrm{Normal}(x^{(\ell-1)},\Sigma) + \mathrm{Normal}(0,\Sigma_\mathrm{w}) \bigr)
  \nonumber \\  &
     \phantom{\eqcom{d}= m^{-1}
     	\Bigl[}
     + \mathrm{Normal}(0,\Sigma_\mathrm{sum}) \Bigr] + \mathrm{Normal}(0,\Sigma_\mathrm{spl})
  \nonumber \\  &
     \small\eqcom{d}=
     m^{-1} \Bigl(\mathrm{Normal}( m W^{(\ell)} x^{(\ell -1)}, m W^{(\ell)}\Sigma ( W^{(\ell)})^\intercal)
  \nonumber \\  &
     + \mathrm{Normal}(0,m\Sigma_\text{w}) + \mathrm{Normal}(0,\Sigma_\mathrm{sum}) \Bigr) + \mathrm{Normal}(0,\Sigma_\mathrm{spl})
  \nonumber \\  &
     \eqcom{d}=\mathrm{Normal}\Bigl( W^{(\ell)}x^{(\ell-1)},
  \nonumber \\  &
     \qquad\qquad
     {m}^{-1} ( W^{(\ell)}\Sigma \bigl( W^{(\ell)}\bigr)^\intercal + \Sigma_\text{w} + {m}^{-1}\Sigma_\mathrm{sum} )
     + \Sigma_\mathrm{spl} \Bigr)
     \nonumber
     .
\end{align}
The random variable $\mathbf{P}^{(\ell)}$ is then channeled through the activation function, which subsequently adds another noise term.
The resulting activation is the random variable $\mathbf{A}^{(\ell)}$
\begin{align}
   &
  \mathbf{A}^{(\ell)}
  \mid
  x^{(\ell-1)} \\ &
  \eqcom{d}= \sigma^{(\ell)} ( \mathbf{P}^{(\ell)} ) + \mathrm{Normal}(0,\Sigma_\text{a}) \\  &
  \eqcom{d}= D^{(\ell)} W^{(\ell)}x^{(\ell-1)}                                        \\  &
     + \mathrm{Normal}
     \Bigl( 0,
     \frac{ D^{(\ell)}W^{(\ell)}\Sigma\bigl( D^{(\ell)}W^{(\ell)}\bigr)^\intercal }{m}
     +
     \frac{ D^{(\ell)} \Sigma_\text{w} \bigl(D^{(\ell)}\bigr)^\intercal }{m}
  \\  &
     \phantom{+ \mathrm{Normal}
       \bigl( 0,}
     +
     \frac{ D^{(\ell)} \Sigma_\mathrm{sum} \bigl(D^{(\ell)}\bigr)^\intercal }{m^2}
     +
     D^{(\ell)} \Sigma_\mathrm{spl} \bigl(D^{(\ell)}\bigr)^\intercal
     +
     \Sigma_\mathrm{a}
     \Bigr)
  \\  &
     \eqcom{d}= \mathrm{Normal}( \Phi^\mathrm{NN}_\ell (x^{(\ell-1)}) , T_m^{(\ell)}( \Sigma ) )
\end{align}

As we can see, instead of
\begin{equation}
  T^{(\ell)}( \Sigma )
  :=
  D^{(\ell)} W^{(\ell)} \Sigma (D^{(\ell)} W^{(\ell)} )^\intercal
  +
  D^{(\ell)} \Sigma_{\mathrm{w}}^{(\ell)} (D^{(\ell)})^\intercal
  +
  \Sigma_a^{(\ell)}
\end{equation}
we have
\begin{align}
  T^{(\ell)}_m( \Sigma )
   &
  :=
  {m}^{-1} D^{(\ell)}W^{(\ell)}\Sigma \bigl( D^{(\ell)}W^{(\ell)}\bigr)^\intercal
  \\  &
     \phantom{:=}
     +
     {m}^{-1}D^{(\ell)} \Sigma_\text{w} \bigl(D^{(\ell)}\bigr)^\intercal
     +
     {m}^{-2}D^{(\ell)} \Sigma_\mathrm{sum} \bigl(D^{(\ell)}\bigr)^\intercal
  \\  &
     \phantom{:=}
     +
     D^{(\ell)} \Sigma_\mathrm{spl} \bigl(D^{(\ell)}\bigr)^\intercal
     +
     \Sigma_\mathrm{a}.
\end{align}

\subsubsection{Proof of \Cref{cor:Results__DesignB_Limit_for_symmetric_linear_ONNs}}

As \Cref{cor:Results__DesignB_Limit_for_symmetric_linear_ONNs} is the Design B analog of \Cref{cor:Limit_for_symmetric_linear_ONNs}, the proofs are similar.
The first expression in \Cref{cor:Results__DesignB_Limit_for_symmetric_linear_ONNs} is again immediate from expansion.
For the limit we use the same Banach fixpoint argument, where only the variables have to be exchanged.
The following executes these steps.

Assume again that for all $\ell \in \mathbb{N}$ $D^{(\ell)}=D$ and $W^{(\ell)}=W$ as well as $\Sigma_\mathrm{a}^{(\ell)} = \Sigma_\mathrm{a}$ and $\Sigma_\mathrm{w}^{(\ell)} = \Sigma_\mathrm{w}$.
Let $T_m:=T_m^{(\ell)}$ be the common map under those conditions.

Recall \eqref{eq:Results__contracting_prop}.
In the setting of \Cref{thm:Results__DesignB_recusive_structure_of_covariances}, that is $X_1=\Sigma_{\mathrm{m}}$ and $X_{\ell +1}= T_m(X_\ell)$, the so-defined sequence converges if $\|D\|_F \|W\|_F <\sqrt{m}$ (see also below \eqref{eq:appendix__tensor_product_goes_to_zero_m}) due to \eqref{eq:Results__contracting_prop} and the Banach fixed point theorem \cite{banach1922operations}.
We therefore let the unique fixed point be
\begin{align}
  \lim_{\ell\rightarrow \infty} X_\ell = X_\star.
\end{align}
We can write the fixed point equation $T_m(X_\star)=X_\star$ as
\begin{align}
   &
  X_\star - m^{-1}(DW) X_\star (DW)^\intercal
  \nonumber \\  &
     =m^{-1} D \Sigma_\mathrm{w} D^\intercal + m^{-1}\Sigma_\mathrm{a} + m^{-2}D \Sigma_\mathrm{sum} D^\intercal + D \Sigma_\mathrm{spl} D^\intercal
     ,
\end{align}
and further write it as
\begin{align}
   &
  \mathrm{vec} (X_\star) - m^{-1}((DW) \otimes (DW)) \mathrm{vec}(X_\star)
  \\
   &
  =
  \mathrm{vec} \bigl( m^{-1}D \Sigma_\mathrm{w} D^\intercal + m^{-1}\Sigma_\mathrm{a} + m^{-2}D \Sigma_\mathrm{sum} D^\intercal
  \\  &
     \phantom{=\mathrm{vec}\bigl(}
     + D \Sigma_\mathrm{spl} D^\intercal\bigr)
     .
\end{align}
Here, $\otimes$ denotes again the Kronecker product and $\mathrm{vec}$ the vectorization of a matrix.
Applying the vectorization trick as in the proof of \Cref{cor:Limit_for_symmetric_linear_ONNs} allows us to write the solution to the fixed point equation as
\begin{align}
   &
  \mathrm{vec} (X_\star)
  \label{eq:Results__solution_fixed point equation_before_neumann_DesignB}
  \\  &
     =
     \bigl( I - (m^{-1}(DW)^{\otimes 2}) \bigr)^{-1}
  \nonumber \\  &
     \phantom{=}
     \mathrm{vec} \bigl( m^{-1}D \Sigma_\mathrm{w} D^\intercal + m^{-1}\Sigma_\mathrm{a} + m^{-2} D \Sigma_\mathrm{sum} D^\intercal
  \\  &
     \phantom{=\mathrm{vec}\bigl(}
     + D \Sigma_\mathrm{spl} D^\intercal\bigr)
     .
\end{align}
Again, $(DW)^{\otimes 2}$ denotes $(DW)\otimes (DW)$.

Formally we rewrite the inverse in \eqref{eq:Results__solution_fixed point equation_before_neumann_DesignB} in terms of a von Neumann series
\begin{align}
  ( I - (m^{-1}(DW)^{\otimes 2}) )^{-1} = \sum_{n=0}^\infty (m^{-1}(DW)^{\otimes 2})^n.
  \label{eq:Results__neumann_in_fixedpointeq_m}
\end{align}
This is again only justified if
\begin{align}
  \| (m^{-1}(DW)^{\otimes 2})^n \|_F \rightarrow 0.
  \label{eq:Results__tensor_product_goes_to_zero_m}
\end{align}
By submultiplicativity it holds that
\begin{align}
  \| (m^{-1}(DW)^{\otimes 2})^n \|_F \leq \| m^{-1} (DW)^{\otimes 2} \|_F^n.
  \label{eq:appendix__tensor_product_goes_to_zero_m}
\end{align}
For the Kronecker product it holds that $\mathrm{tr}(A\otimes B)=\mathrm{tr}(A) \allowbreak \mathrm{tr}(B)$ and thus $\| (DW)^{\otimes 2} \|_F = \mathrm{tr}( (DW)^\ast (DW) ) = \| DW \|_F^2$ by definition of the Frobenius norm.
Therefore, by our assumption of $\|D\|_F \|W\|_F <\sqrt{m}$, condition \eqref{eq:Results__tensor_product_goes_to_zero_m} holds and \eqref{eq:Results__neumann_in_fixedpointeq_m} is valid.
To simplify the notation we let $\Sigma_\mathrm{sum}=\Sigma_\mathrm{spl}=0$, leading to the representation of $X_\ast$ as
\begin{align}
  \mathrm{vec} (X_\star)
  =
   &
  \sum_{n=0}^\infty
  m^{-n}((DW)^{\otimes 2})^n \mathrm{vec}( m^{-1} D \Sigma_\mathrm{w} D^\intercal + \Sigma_\mathrm{a} )	.
  \label{eq:Results__solution_fixed point equation_with_neumann_DesignB}
\end{align}
Returning to the matrix notation we have
\begin{align}
  X_\star =
  \sum_{n=0}^\infty m^{-n} (DW)^n ( m^{-1}D \Sigma_\mathrm{w} D^\intercal + \Sigma_\mathrm{a}) ((DW)^n)^\intercal.
  \label{eq:appendix__solution_fixpoint}
\end{align}
That is it.
\QuodEratDemonstrandum

\end{document}